\documentclass[letterpaper]{article}

\usepackage{natbib,alifeconf}  

\usepackage[dvipsnames,table,xcdraw]{xcolor}
\usepackage{graphicx}

\usepackage{adjustbox}
\usepackage[noEnd=true,commentColor=black]{algpseudocodex}
\usepackage[plain]{algorithm}
\usepackage{amsmath,amssymb,amsthm,amsfonts}
\usepackage{annotate-equations}
\usepackage{array}
\usepackage{bibunits}
\usepackage{bold-extra}
\usepackage{booktabs}
\usepackage{cancel}
\usepackage{caption}
\usepackage{circledsteps}
\usepackage{cite}
\usepackage{csvsimple}
\usepackage{etoolbox}
\usepackage{float}
\newfloat{algorithm}{t}{lop} 
\usepackage[margin=1in]{geometry}
\usepackage{hyperref}
\usepackage{import}
\usepackage{mathtools}
\usepackage{orcidlink}
\usepackage{physics}
\usepackage{placeins}
\usepackage[moderate]{savetrees}
\usepackage{siunitx}
\usepackage{stix}
\usepackage{stmaryrd}
\usepackage{subcaption}
\usepackage{svg}
\usepackage{tabularx}
\usepackage{textcomp}
\usepackage{tikz}
\usetikzlibrary{arrows.meta}
\usepackage[most]{tcolorbox}
\usepackage{xparse}
\usepackage{pythonhighlight}
\usepackage{empheq}
\usepackage{cleveref}
\usepackage[compact]{titlesec}

\definecolor{seabornBrown}{RGB}{134, 86, 75}
\definecolor{seabornPink}{RGB}{255, 119, 193}
\definecolor{seabornPinkDark}{RGB}{152, 34, 116}

\newcolumntype{Y}{>{\centering\arraybackslash}X}

\newtcolorbox{mybox}{
enhanced,
boxrule=0pt,frame hidden,
borderline west={4pt}{0pt}{green!50!black},
colback=green!30!gray!15,
sharp corners,
parbox=false
}

\makeatletter
\def\mathcolor#1#{\@mathcolor{#1}}
\def\@mathcolor#1#2#3{%
  \protect\leavevmode
  \begingroup\color#1{#2}#3\endgroup
}
\makeatother

\makeatletter
\DeclareRobustCommand{\ctau}{{
  \mathpalette\cap@greek\tau
}}
\DeclareRobustCommand{\csigma}{{
  \mathpalette\cap@greek\sigma
}}
\newcommand{\cap@greek}[2]{%
  \begingroup
  \sbox\z@{$#1t$}
  \resizebox{!}{\ht\z@}{$\m@th#1#2$}
  \endgroup
}
\makeatother

\let\mycheckmark\checkmark
\renewcommand{\checkmark}{\textcolor{ForestGreen}{\mycheckmark}}

\theoremstyle{definition}

\makeatletter
\@addtoreset{proofpart}{theorem}
\@addtoreset{proofpart}{lemma}
\@addtoreset{proofpart}{corollary}
\makeatother

\hypersetup{breaklinks=true}

\renewcommand{\eqnhighlightheight}{\mathstrut}

\newcommand*{\colorboxed}{}
\def\colorboxed#1#{%
  \colorboxedAux{#1}%
}
\newcommand*{\colorboxedAux}[3]{%
  \begingroup
    \setlength\fboxrule{1pt}
    \colorlet{cb@saved}{.}%
    \color#1{#2}%
    \boxed{%
      \color{cb@saved}%
      #3%
    }%
  \endgroup
}

\makeatletter
\renewcommand{\boxed}[1]{\text{\fboxsep=.2em\fbox{\m@th$\displaystyle#1$}}}
\makeatother

\let\mythealgorithm\thealgorithm

\makeatletter
\newlength{\comment@width}

\renewcommand{\Comment}[1]{%
  \sbox0{#1}
  \ifdim\wd0>\comment@width
    \setlength{\comment@width}{\wd0}%
  \fi
  \ifcsname comment@\arabic{algorithm}@width\endcsname
    \algorithmiccomment{\makebox[\csname comment@\mythealgorithm @width\endcsname][l]{#1}}%
  \else
    \algorithmiccomment{#1}%
  \fi
}
\AtBeginEnvironment{algorithmic}{\setlength{\comment@width}{0pt}}
\AtEndEnvironment{algorithmic}{%
  \immediate\write\@auxout{%
    \string\algcommentwidth{\mythealgorithm}{\the\comment@width}%
  }%
}
\newcommand{\algcommentwidth}[2]{%
  \global\@namedef{comment@#1@width}{#2}%
}
\makeatother

\MakeRobust{\Call}
\makeatletter
\def\algbackskip{\hskip-\ALG@thistlm}
\makeatother

\lstnewenvironment{mypython}[1][]{\lstset{style=mypython,#1}}{}

\definecolor{darkred}{HTML}{E32B60}

\definecolor{codegreen}{rgb}{0,0.6,0}
\definecolor{codegray}{rgb}{0.5,0.5,0.5}
\definecolor{codepurple}{rgb}{0.58,0,0.82}
\definecolor{backcolour}{rgb}{0.95,0.95,0.92}

\lstdefinestyle{mystyle}{
    backgroundcolor=\color{backcolour},
    commentstyle=\color{codegreen},
    keywordstyle=\color{magenta},
    numberstyle=\tiny\color{codegray},
    stringstyle=\color{codepurple},
    basicstyle=\ttfamily\footnotesize,
    breakatwhitespace=false,
    breaklines=true,
    captionpos=t,
    keepspaces=true,
    numbers=left,
    numbersep=5pt,
    showspaces=false,
    showstringspaces=false,
    showtabs=false,
    tabsize=2
}

\DeclareCaptionFormat{listing}{\rule{\dimexpr\textwidth\relax}{0.4pt}\par\vskip1pt#1#2#3\par\vspace{-5pt}\rule{\dimexpr\textwidth\relax}{0.4pt}}
\makeatletter
\let\pragma@iinput=\@iinput
\def\@iinput#1{\xdef\@pragmafile{#1}\pragma@iinput{#1} }
\def\@pragmafile{default}
\def\pragmaonce{%
   \csname pragma@\@pragmafile\endcsname
   \global\expandafter\let \csname pragma@\@pragmafile\endcsname =  
}
\makeatother

\ifdefined\mydraft
\mydraft
\fi

\defaultbibliography{bibl}
\defaultbibliographystyle{apalike-ejor}

\begin{document}

\title{A Scalable Trie Building Algorithm for High-Throughput Phyloanalysis of Wafer-Scale Digital Evolution Experiments}

\author{
    Vivaan Singhvi\orcidlink{0009-0005-8628-8221}$^{1,5,6,\dagger}$,\ %
    Joey Wagner\orcidlink{0009-0000-6141-976X}$^{6,9}$,\ %
    Emily Dolson\orcidlink{0000-0001-8616-4898}$^{7,8,9}$,\ %
    Luis Zaman\orcidlink{0000-0001-6838-7385}$^{2,3,5}$, \and
    Matthew Andres Moreno\orcidlink{0000-0003-4726-4479}$^{2,3,4,5,\ddagger}$ \\
    \mbox{}\\
    $^1$Michigan Research and Discovery Scholars
    $^2$Department of Ecology and Evolutionary Biology \\
    $^3$Center for the Study of Complex Systems
    $^4$MIDAS
    $^5$University of Michigan, Ann Arbor, United States \\
    $^6$Professorial Assistantship Program
    $^7$Department of Computer Science and Engineering \\
    $^8$Program in Ecology, Evolution, and Behavior
    $^9$Michigan State University, East Lansing, United States \\
    $^\dagger$\texttt{vsinghvi@umich.edu} $^\ddagger$\texttt{morenoma@umich.edu}
}

\maketitle

\begin{bibunit}

\begin{abstract}
Agent-based simulation platforms play a key role in enabling fast-to-run evolution experiments that can be precisely controlled and observed in detail.
Availability of high-resolution snapshots of lineage ancestries from digital experiments, in particular, is key to investigations of evolvability and open-ended evolution, as well as in providing a validation testbed for bioinformatics method development.
Ongoing advances in AI/ML hardware accelerator devices, such as the 850,000-processor Cerebras Wafer-Scale Engine (WSE), are poised to broaden the scope of evolutionary questions that can be investigated \textit{in silico}.
However, constraints in memory capacity and locality characteristic of these systems introduce difficulties in exhaustively tracking phylogenies at runtime.
To overcome these challenges, recent work on hereditary stratigraphy algorithms has developed space-efficient genetic markers to facilitate fully decentralized estimation of relatedness among digital organisms.
However, in existing work, compute time to reconstruct phylogenies from these genetic markers has proven a limiting factor in achieving large-scale phyloanalyses.
Here, we detail an improved trie-building algorithm designed to produce reconstructions equivalent to existing approaches.
For modestly-sized 10,000-tip trees, the proposed approach achieves a 300-fold speedup versus existing state-of-the-art.
Finally, using 1 billion genome datasets drawn from WSE simulations encompassing 954 trillion replication events, we report a pair of large-scale phylogeny reconstruction trials, achieving end-to-end reconstruction times of 2.6 and 2.9 hours.
In substantially improving reconstruction scaling and throughput, presented work establishes a key foundation to enable powerful high-throughput phyloanalysis techniques in large-scale digital evolution experiments.
\end{abstract}

\section{Introduction} \label{sec:introduction}

Key aspects of the study of evolution, whether biological or digital, revolve around understanding the flow of genetic material among large populations of organisms.
As such, phylogenetic analyses assessing ancestry trees representing organisms' evolutionary histories are a core tool in evolutionary biology.

In biology, phylogeny estimations are typically reconstructed through \textit{post hoc} analysis of genetic similarities among organisms.
In contrast, direct, exact tracking at runtime is typical in \textit{in silico} experiments.
However, in memory-constrained parallel and distributed computing contexts, \textit{post hoc} reconstruction approaches can become advantageous owing to runtime synchronization and storage costs of direct tracking.

Akin to biological studies, efficacy of phylogenetic analysis in such digital experiments hinges on fast, accurate methods to estimate ancestry trees from genome data.
In this work, we present a novel trie-building algorithm that greatly reduces compute time necessary to reconstruct phylogenies from special-purpose markers on digital genomes, while producing results equivalent to a naive approach.

\subsection{Applications of Phylogenetic Analysis}

Phylogenetic analyses provide key means to characterize and quantify a broad array of evolutionary processes.
Classically, these analyses have been applied to investigation of species-level macroevolutionary dynamics revolving around speciation and extinction rates; however, population- and organism-level dynamics can also be inferred, such as the spread of beneficial mutations within a population or fitness parameters like growth rate and probability of survival \citep{genthon2023cell, levy2015quantitative, stadler2013recovering}.
Phylogenetic analysis is also crucial in the field of epidemiology, playing a key role in informing public health interventions.
In this context, phylogenetic methods can be used to determine transmission history, pinpointing where and how chains of infection unfold \citep{wang2020role}.
In this vein, phylogenies are also key in assessing the prevalence of  ``super-spreader'' dynamics wherein disease spread is driven by a small set of high-risk individuals \citep{colijn2014phylogenetic}.



\subsection{Phylogenies and Digital Evolution} \label{sec:introduction:digital}

In some cases, aspects of biological evolution can be difficult or infeasible to observe on human timescales; laboratory experiments may take years, or even decades, to complete \citep{wiser2013long,Stroud2025}.
By simulating the behavior of a population, some experiments can instead be conducted digitally --- often completing in a fraction of the time.
Digital experiments can model key characteristics of biological populations, such as variation, natural selection, ecological interactions, spatial distribution, and more \citep{dolson2021digital,haller2023slim}.
As such, conclusions from digital evolution experiments can contribute meaningfully to understanding biology \citep{pennock2007models}.

In digital evolution contexts, phylogenies have likewise proven valuable.
In application-oriented contexts, phylogeny-based biodiversity metrics have been shown as predictive of solution quality outcomes for evolutionary computation-based optimization \citep{hernandez2022phylogenetic}.
Phylogeny-based methods can also be applied to characterize more general aspects of ecology, spatial structure, and selection pressure within \textit{in silico} populations \citep{moreno2023toward}.

Digital evolution approaches can also serve as a testbed to assess bioinformatics methodologies.
The Aevol\_4b system, for instance, uses a genetic system corresponding to that of DNA, allowing any genetic information to be processed using methods directly from bioinformatics \citep{daudey2024aevol}.
Likewise, population genetics work often incorporates SLiM, which supports sophisticated continuous-space modeling of single- and multi-species systems \citep{haller2023slim}.

Given the programmatic observability of digital simulations, digital evolution platforms typically incorporate direct tracking methods that record lineage ancestry as the simulation runs.
General-purpose phylogeny-tracking libraries exist for this purpose \citep{dolson2024phylotrack}, although many platforms simply incorporate bespoke implementations into their own software \citep{ofria2004avida}.

\subsection{Scaling Up Digital Evolution Experiments} \label{sec:introduction:distributed}

To achieve large-scale digital evolution experiments, it is necessary to move from a single-processor system to a more distributed approach with many computing units \citep{moreno2024trackable}.
In large-scale, many-processor simulations, however, challenges arise in managing a comprehensive record of ancestry.
To control memory use, it is typically necessary to trim away records of extinct lineages when performing direct tracking.
Detecting extinctions, however, introduces implementation complexity and overhead costs when lineage histories span across multiple processors.
Exhaustive tracking is also sensitive to data loss from crashed hardware or dropped messages, which has been highlighted as a key consideration in achieving very large-scale artificial life systems \citep{ackley2016indefinite,ackley2014indefinitely}.

Challenges associated with comprehensive tracking are especially acute in specialized hardware accelerator devices, which represent a promising emerging direction in high-performance computing \citep{emani2024democratizing}.
In incorporating thousands of processor cores per device, these hardware architectures impose trade-offs in memory capacity limitations and data locality restrictions that limit the feasibility of comprehensive tracking.
In such contexts, reconstruction-based approaches can provide an attractive balance between data fidelity and data collection overhead.

\subsection{Hereditary Stratigraphy} \label{sec:introduction:hstrat}

\begin{figure*}[h]

\centering
\begin{minipage}{0.55\textwidth}

\begin{minipage}{0.41\linewidth}
\centering
\includegraphics[height=1.1in]{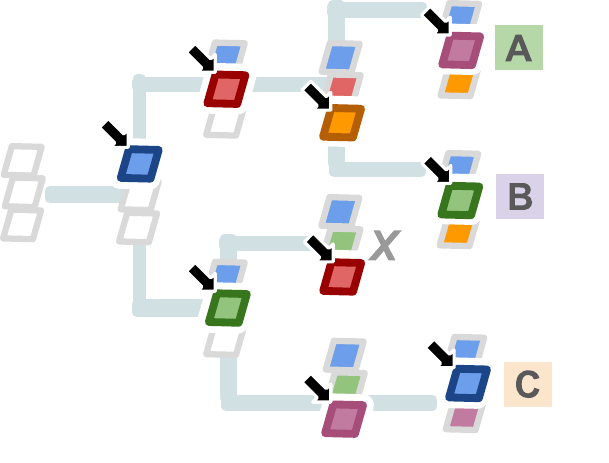}
\subcaption{evolve}
\label{fig:hstratschematic:evolve}
\end{minipage}%
\vrule
\centering
\begin{minipage}{0.18\linewidth}
~
\includegraphics[height=1.1in]{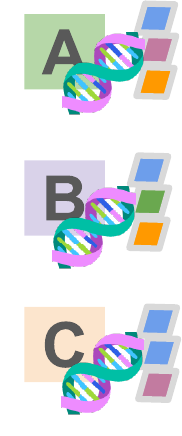}
\subcaption{sample}
\label{fig:hstratschematic:sample}
\end{minipage}%
\vrule
\begin{minipage}{0.41\linewidth}
\centering
\includegraphics[height=1.1in]{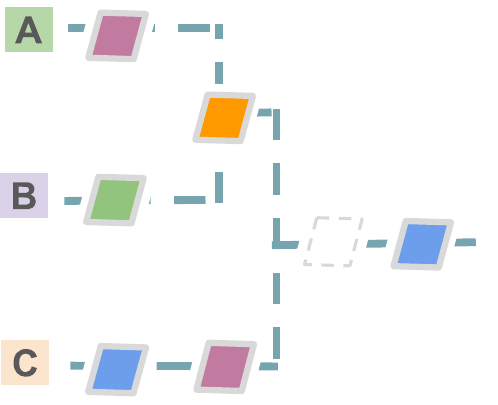}
\subcaption{reconstruct}
\label{fig:hstratschematic:reconstruct}
\end{minipage}
\end{minipage}%
~~
\begin{minipage}{0.43\textwidth}
\caption{%
\textbf{Overview of hereditary stratigraphy.}
\small
At runtime, genomes are annotated with randomly-generated heritable markers (panel \ref{fig:hstratschematic:evolve}).
To maintain fixed-memory footprint, some markers are overwritten.
Genomes of interest are sampled at runtime and from end state (panel \ref{fig:hstratschematic:sample}).
Decoded genome markers enable estimation of evolutionary relatedness (panel \ref{fig:hstratschematic:reconstruct}), subject to error from marker-value collisions and discarded markers.
}
\label{fig:hstratschematic}
\end{minipage}
\end{figure*}

Under controlled conditions, such as laboratory experiments or evolution simulations, genetic material may be engineered to facilitate the accuracy and efficiency of estimating phylogenetic relatedness \citep{li2024reconstructing,ackley2023robust}.%
\footnote{Notably, Ackley has applied barcoding approaches to track recent ancestry among emergent replicators in a distributed fabric-computing context.}
Work developing hereditary stratigraphy (``hstrat'') methods seeks to operate analogously, providing techniques to organize genetic material in digital organisms that maximize reconstruction quality while minimizing memory footprint \citep{moreno2022hereditary}.
Hereditary stratigraphy components can be bundled with agent genomes in a manner akin to non-coding DNA (i.e., neutral with respect to agent traits and fitness), enabling generalizability across a wide variety of agent models.

Hereditary stratigraphy associates each generation along each lineage with an identifying ``fingerprint'' marker, referred to as a differentia.
On birth, each offspring receives a new differentia value and appends it to an inherited chronological record of past values --- corresponding to earlier generations along its lineage.
Under this scheme, mismatching differentia can be used to delimit the end of common ancestry between two organisms.
Figure \ref{fig:hstratschematic} summarizes this approach.

To save space, differentiae may be pruned away --- although, at the cost of reducing precision in inferring relatedness.
Using fewer bits per differentia can also provide many-fold memory savings; single bits or single bytes are appropriate for most use cases.

While inferring relatedness from biological sequence data can be a highly challenging and computationally-intensive problem \citep{miller2010creating},
the structured marker data used in hereditary stratigraphy somewhat ameliorates this challenge by allowing phylogeny reconstruction to be approached as a trie-building problem of identifying common string prefixes \citep{delabriandais1959file,moreno2024analysis}.
However, the presence of missing data due to some differentia being dropped to save memory complicates matters.

In the context of trie-building, missing marker time points (possessed by only a subset of organisms) effectively act as ``wildcard'' characters in prefix matching operations.
Therefore, placing an organism on a trie requires evaluating diverging string paths beyond the wildcard to identify further matches.
Given the likelihood of differentia value collisions for small differentia sizes (e.g., 1 bit), identifying the best-matching path after a wildcard value can require looking ahead several consecutive markers.
Furthermore, where consecutive wildcard values are encountered, the number of possible paths that must be explored can grow exponentially.

Although previous work has investigated the quality of phylogenies constructed from hereditary stratigraphy data using trie-based approaches \citep{moreno2025testing}, the computational intensity of the naive wildcard-matching approach has limited the scale of phylogenetic reconstructions investigated and restricted experimental throughput for smaller reconstructions.
Given the objective of hereditary stratigraphy methodology to facilitate studying very large-scale digital evolution experiments, achieving reconstruction efficiency sufficient for large-scale phyloanalysis is critical to the overall utility of the methodology in enabling observable experiments.

In this work, we describe an algorithm for efficient trie reconstruction in the face of missing data, and explore its performance characteristics.
The following section introduces our proposed ``shortcut'' algorithm for the trie building approach explored in this paper.
We then detail methods and results for benchmark trials assessing empirical scaling behavior and performance on large-scale billion-genome workloads.

\section{Proposed Algorithm} \label{sec:algorithm}

In this section, we overview the existing naive approach used for constructing prefix tries of hereditary stratigraphy markers, and describe the proposed shortcut-based approach developed in this work.
Notation describes hereditary stratigraphy markers as a pair of two properties: \textit{rank} $r$, referring to the generation at which a marker was generated, and \textit{differentia} $d$, referring to the randomly-generated value distinguishing a given marker from others generated at that generation.

\subsection{Naive Trie-Building Algorithm} \label{sec:algorithm:naive}

The naive algorithm treats each organism's sequence of stored rank-differentia pairs as a string: each differentia $d$ constitutes a character at the rank $r$-th string position.
Differentia values missing due to having been overwritten are considered as wildcard characters.

For two rank-differentia strings, common ancestry corresponds to a shared prefix of string characters inherited from their most recent common ancestor (MRCA).
Under this framing, phylogenies correspond to a trie --- a tree-like data structure where each node corresponds to a character in a string, and each path from the root to a leaf node reads out a stored string \citep{fredkin1960trie}.

The first step in naive reconstruction is to sort all organisms in ascending order by number of generations elapsed.
Due to each organism using the same algorithm to determine which markers to overwrite, this order guarantees that any data missing from an organism will also have been discarded by all subsequent organisms.

We initialize an empty trie, consisting of a single root node.
Trie building proceeds by processing organisms one at a time in their sorted order.
For each organism, we iterate through rank-differentia pairs in chronological order.
At each rank-differentia pair, we either continue along the existing trie, branch off to create a new path, or --- in a special case --- address missing information (Algorithm~\ref{alg:old}).

\begin{algorithm}[h]
    \begin{algorithmic}[1]
    \small{
        \Function{ReconstructTree}{organism list $O$}
            \State $T \gets$ an empty tree
            \For {$o \in O$}
                \State $\textsc{TreeInsert}(T,\; o)$
            \EndFor
            \State \Return $T$
        \EndFunction

        \Function{TreeInsert}{tree $T$, organism $o$}
            \State $n \gets$ root of $T$
            \For{$(r,\; d) \in o$} \text{ } $\triangleright$ \textit{successive rank-differentia pairs}
                \State $n \gets \textsc{MostLikelyChild}(n,\; o)$ \text{ } $\triangleright$ \textit{see supplement}
                \If{$\exists c \in \operatorname{children}(n) \text{ s.t.} \; \operatorname{differentia(c) = d \text{ and } \operatorname{rank}(c) = r}$}
                    \State $n \gets c$
                \Else
                    \State $c' \gets \textsc{CreateChild}(n,\; r,\; d)$ \text{ } $\triangleright$ \textit{new child off $n$}
                    \State $n \gets c'$
                \EndIf
            \EndFor
            \State $o$.parent $\gets n$ ~~~ $\triangleright$ \textit{attach organism $o$ to tree}
        \EndFunction
    }
    \end{algorithmic}
    \caption{\textbf{Naive trie-building algorithm.} \small Iteratively builds a trie from organisms' genetic markers. Requires a list of organisms $O$ in ascending order by generations elapsed. Within organisms, genetic markers are stored as a chronological list of rank-differentia $(r, d)$ pairs. \vspace{-1.5em}}
    \label{alg:old}
\end{algorithm}

Missing information is recognized as follows.
Suppose we have reached trie node $n$ with rank $r_1$ and must next process rank-differentia pair $(r_2,\; d)$ from current organism $o$.
If any node with rank $r' < r_2$ exists among children of current node $n$, we know that $r'$ must have been deleted from organism $o$ as its record skips directly from $r_1$ to $r_2$.

To maximize reconstruction accuracy in the case of missing data, we must assess which --- if any --- of $n$'s children our missing datum at $r'$ would have likely corresponded to.
Essentially, when reaching missing information, we must infer that information, which we do by looking ahead in the tree.
Specifically, we search forward for the path with the longest successive streak of differentia matching to current organism $o$ \citep{moreno2024analysis}.
If there is no matching path, we simply branch off node $n$ rather than continuing to traverse through its children.

Note that this approach suffers the cost of doing significant extra work to handle missing information.
In the worst case, there could be an exponential number of matching paths to check, owing to the possibility of nested branch-outs at successive wildcard sites.
These searches are repeated for each organism, regardless of whether or not a similar search was already done.
Therefore, we present an improved algorithm that consolidates this extra work into a single step that never needs to be repeated.

\subsection{Proposed Shortcut Algorithm} \label{sec:algorithm:shortcut}

\begin{figure*}
\begin{minipage}{0.74\linewidth}
\centering
\begin{minipage}{0.32\linewidth}
  \centering
  \includegraphics[width=\linewidth]{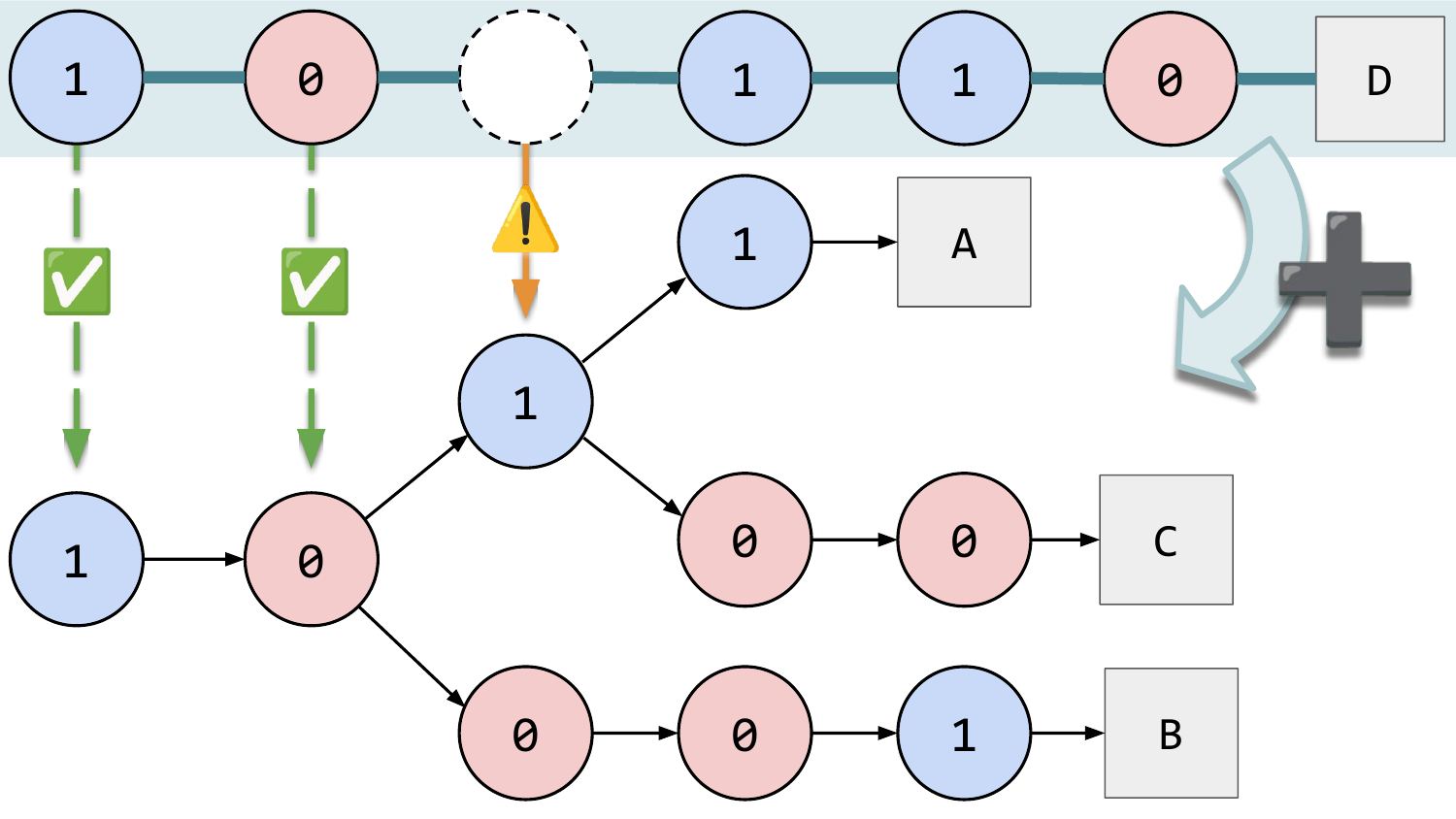}
  \subcaption{Preparing to add $D$}
  \label{fig:shortcut-algo-diagram-1}
\end{minipage}
\begin{minipage}{0.32\linewidth}
  \centering
  \includegraphics[width=\linewidth]{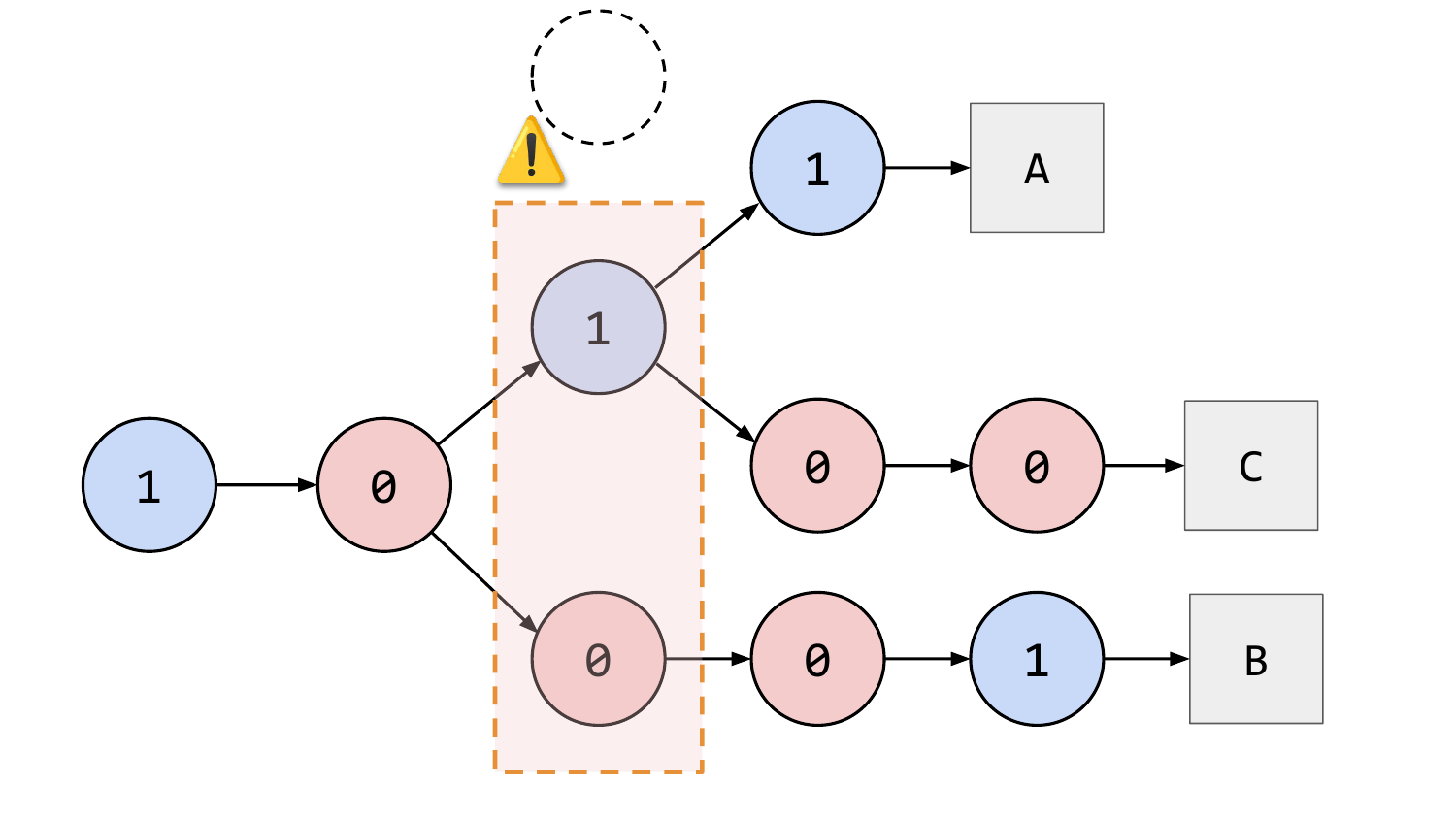}
  \subcaption{Encountering dropped marker}
  \label{fig:shortcut-algo-diagram-2}
\end{minipage}
\begin{minipage}{0.32\linewidth}
  \centering
  \includegraphics[width=\linewidth]{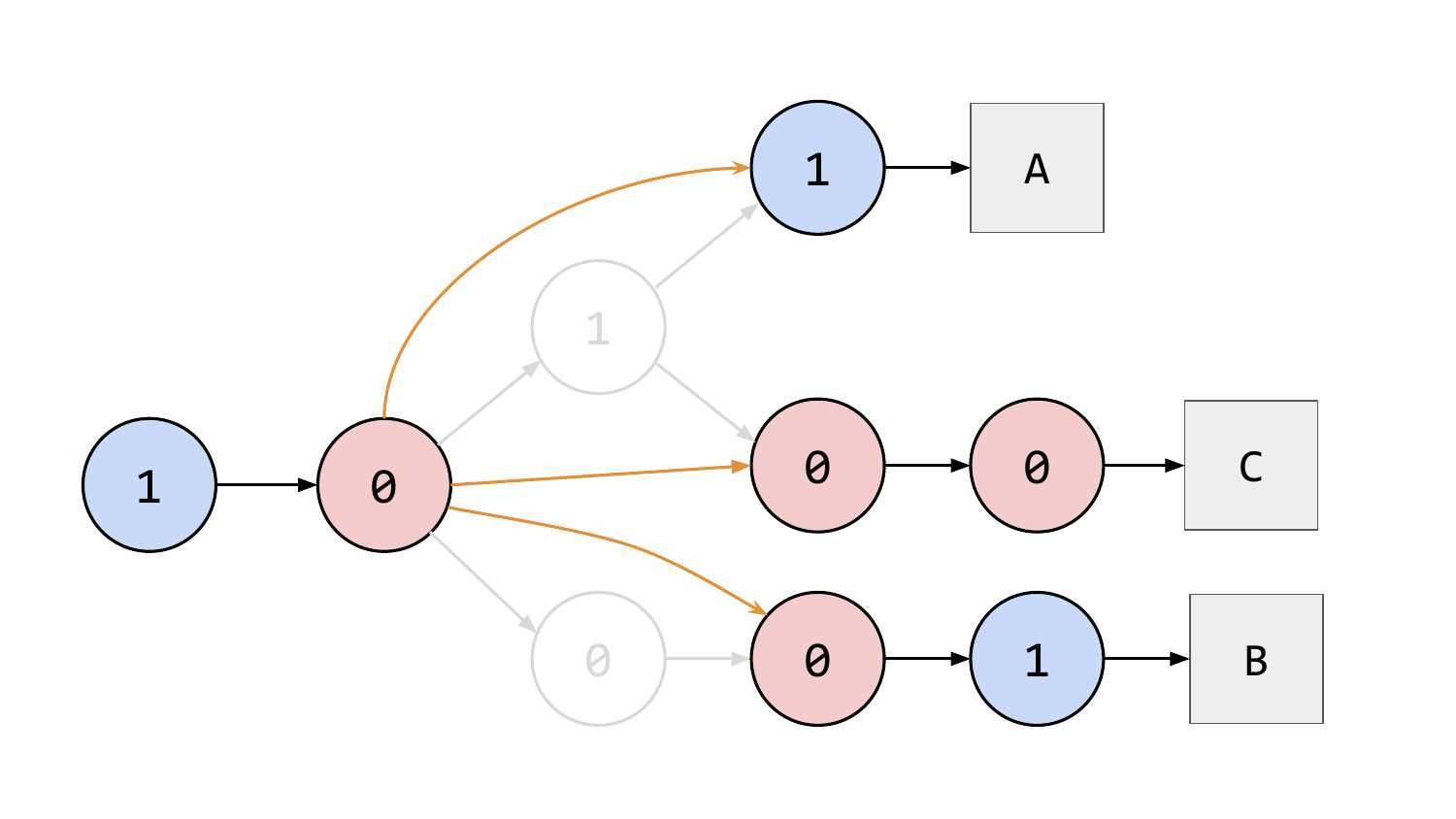}
  \subcaption{Building shortcuts}
  \label{fig:shortcut-algo-diagram-3}
\end{minipage}
\begin{minipage}{\linewidth}\par\end{minipage}
\begin{minipage}{0.32\linewidth}
  \centering
  \includegraphics[width=\linewidth]{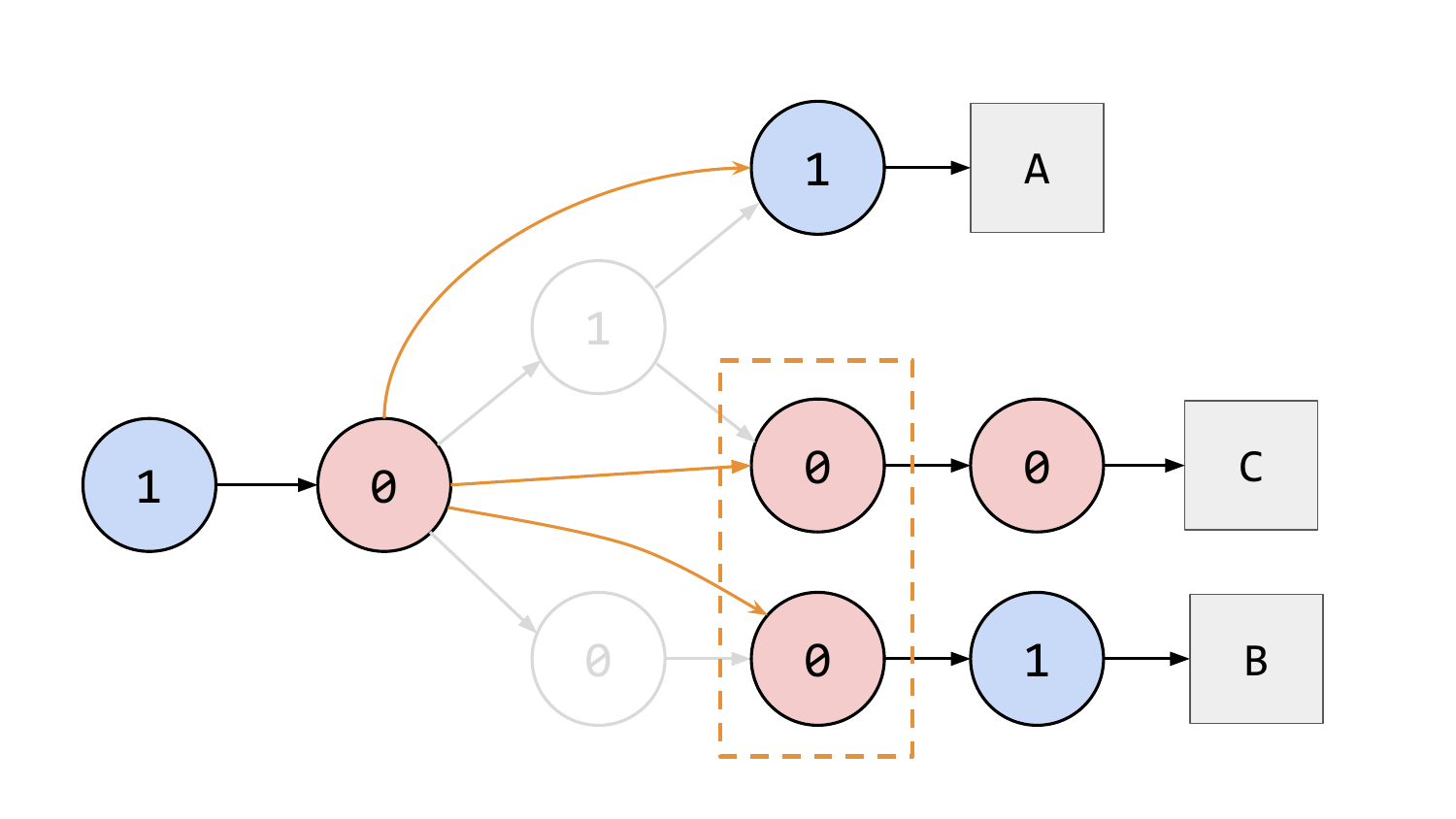}
  \subcaption{Indistinguishable nodes}
  \label{fig:shortcut-algo-diagram-4}
\end{minipage}
\begin{minipage}{0.32\linewidth}
  \centering
  \includegraphics[width=\linewidth]{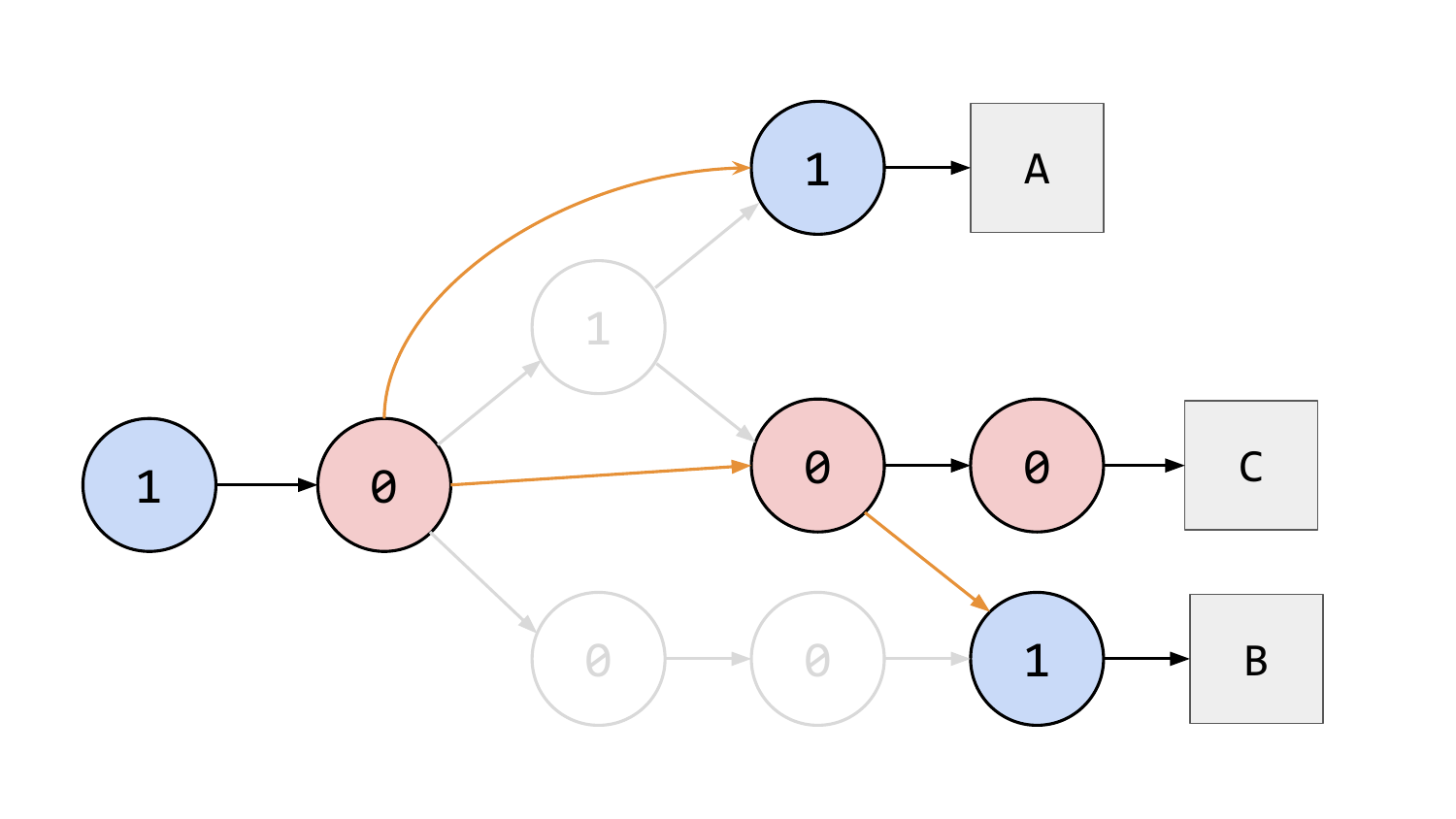}
  \subcaption{Collapsing redundant shortcuts}
  \label{fig:shortcut-algo-diagram-5}
\end{minipage}
\begin{minipage}{0.32\linewidth}
  \centering
  \includegraphics[width=\linewidth]{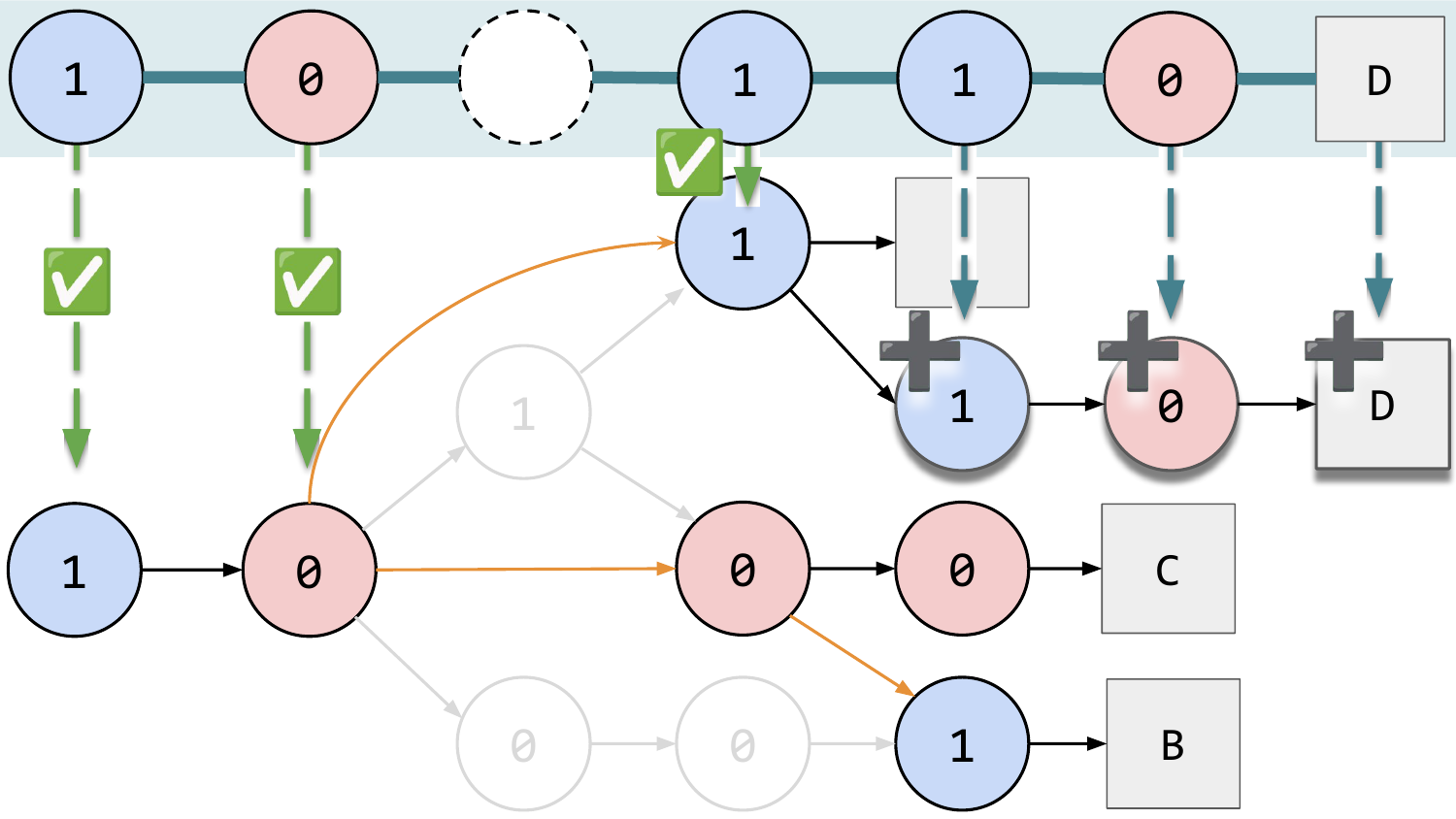}
  \subcaption{Organism $D$ added}
  \label{fig:shortcut-algo-diagram-6}
\end{minipage}
\end{minipage}
\begin{minipage}{0.24\linewidth}
\caption{%
\textbf{Trie-consolidation procedure for proposed shortcut algorithm.}
\small
A dropped hereditary marker is encountered while extending trie with genome $D$ (panel \ref{fig:shortcut-algo-diagram-2}).
All subsequent-added genomes will also have dropped markers at this position, so corresponding trie nodes may be bypassed by ``shortcut'' connections (panel \ref{fig:shortcut-algo-diagram-3}).
Note that bypassed trie structure is retained (``grayed-out'' nodes), so corresponding phylogenetic structure remains when reconstruction is finalized.
In a final step, shortcuts leading to indistinguishable nodes are further consolidated (panels \ref{fig:shortcut-algo-diagram-4} and \ref{fig:shortcut-algo-diagram-5}).
}
\label{fig:algo-diagram}
\end{minipage}
\end{figure*}

We propose a modification of Algorithm~\ref{alg:old} that replaces the $\textsc{MostLikelyChild}$ search in line 9 with a shortcut-building and consolidation step to deal with missing ranks.

Recall that once data from a specific rank is absent, it will not appear in any subsequent organisms.
Hence, trie nodes corresponding to that rank are no longer informative in placing subsequent data.
By building shortcuts bypassing such irrelevant nodes, we can speed up trie traversal for subsequent inserts.

In the shortcut-enabled approach, we detect missing information the same way as before: encountering some trie node $n'$ with a rank $r'$ that falls between the current organism $o$'s last-inserted rank $r_1$ and its currently-inserting rank $r_2$ (i.e., $r_1 < r' < r_2$).

Suppose we detect missing information while stepping from trie node $n$.
To proceed with shortcut construction, we will flesh out a full set $B$ of nearby no-longer-informative nodes.
We build $B$ by collecting all descendants of $n$ with rank $r''$ falling in the gap $r_1 < r'' < r_2$.
We then create shortcut edges that bypass nodes in $B$ --- connecting $n$ directly to the immediate descendants of $B$ with rank $\geq r_2$.

For illustration, see Figure~\ref{fig:algo-diagram}, where we detect a dropped rank (panel \ref{fig:shortcut-algo-diagram-2}) and create shortcut edges trie nodes with that dropped rank (panel \ref{fig:shortcut-algo-diagram-3}).
Edges to bypassed nodes $B$ are ignored in subsequent insert traversals, in favor of added shortcut edges.

One complicating factor may arise in shortcut construction: the introduction of ``indistinguishable nodes,'' where new sibling nodes happen to share the same rank and differentia value (like panel \ref{fig:shortcut-algo-diagram-4}).
This situation can be resolved by arbitrarily choosing one node to discard and transferring its children to the kept node using shortcut edges --- essentially `merging' the duplicates (panel \ref{fig:shortcut-algo-diagram-5}).

Algorithm~\ref{alg:consolidation} outlines steps for the described shortcut construction and consolidation procedure.

 \begin{algorithm}[h]
  \begin{algorithmic}[1]
  \small{
      \Function{ConsolidateTree}{tree $T$, node $n$, rank $r$}
      \State $S \gets \{c \in \operatorname{descendants}(n) \\~~~~~~~~~~~: \operatorname{rank}(c) \ge r \text{ and }  \operatorname{rank}(\operatorname{parent}(c)) < r\}$
      \For{$c$ in $S$}
        \textsc{BuildShortcut}($T$, $n$, $c$)
      \EndFor
      \For{ $(r,\;d) \in \{(\operatorname{rank}(s),\; \operatorname{differentia}(s)) : s \in S \}$}
        \State $S' \gets \{s \in S : \operatorname{rank}(s) = r \text{ and } \operatorname{differentia}(s) = d\}$
        \If {$|S'| = 1$}
          \State \textbf{continue}
        \EndIf
        \State $c^* \gets$ an arbitrary element in $S'$
        \For{$c' \in S' \setminus \{c^*\}$}
          \For{$c''$ in $\operatorname{children}(c')$}
            \State \textsc{BuildShortcut}($T$, $c^*$, $c''$)
          \EndFor
          \State \textsc{RemoveShortcut}($T$, $n$, $c'$)
        \EndFor
      \EndFor
    \EndFunction
    \Function{BuildShortcut}{tree $T$, node $n$, node $c$}
      \State $\operatorname{edges}(T) \gets \operatorname{edges}(T) \cup \{(n,\; c)\}$
      \State $\operatorname{edges}(T)[(n,\; c)]\text{.is\_shortcut} \gets \textsc{True}$
    \EndFunction
    \Function{RemoveShortcut}{tree $T$, node $n$, node $c$}
        \State $\operatorname{edges}(T) \gets \operatorname{edges}(T) \setminus \{(n,\; c)\}$
    \EndFunction
  }
  \end{algorithmic}
  \caption{\textbf{Shortcut-building and consolidation procedure.} \small Builds shortcuts from a given node $n$ and collapses ``indistinguishable'' sibling sets introduced by those shortcuts. \vspace{-1.5em}}
  \label{alg:consolidation}
\end{algorithm}

After insertion of the final organism into our trie, all that remains is to export a final phylogeny output.
At this stage, we ignore shortcut edges.
Crucially, while shortcut edges are useful for efficient traversal during insertions, they do not represent meaningful lineage relationships.
Instead, we use original trie edges as the basis for phylogeny output --- thereby preserving the full detail and granularity of the naive approach.
It is therefore necessary to retain records of the original trie structure, represented by light gray nodes and edges in Figure~\ref{fig:algo-diagram}.
Details about algorithm implementation and more detailed discussion of naive equivalency are provided in supplemental material \citep{supplemental}.

\section{Methods} \label{sec:methods}

In this section, we detail the methodology for benchmark and validation experiments used to assess the performance of the proposed shortcut-enabled algorithm.

\subsection{Generation of Benchmark Data}

Across our benchmark experiments, we needed a large-scale sample of genomes to perform reconstructions on.
For this purpose, we performed simulations using the Cerebras CS-2
Wafer-Scale Engine, an 850,000-core hardware accelerator device \citep{lie2023cerebras}.
Processor elements (PEs) in this architecture are arranged in a grid lattice, with each processor able to execute independently and communicate with neighboring PEs through message-passing.

To conduct these simulations, we used an extension of the island-model genetic algorithm framework presented in \citep{moreno2024trackable}.
This framework organizes simulated organisms into independent subpopulations hosted on each PE.
Synchronous tournament selection of size 2 is applied within each PE.
Between generations, agents are migrated between PEs asynchronously.


A simple genome model was used, comprised of a single floating point value representing fitness.
Reproduction occurred asexually, with this value subject to a normally-distributed mutation with 30\% probability.
(In the case of the purifying-regime treatment, the negative absolute value of a sampled mutation delta was applied.)
In tournament competitions to select parents for the subsequent generation, the parent was selected according to the higher fitness value with 10\% probability and was otherwise selected randomly.
Genomes were augmented with hstrat annotations comprising 64 single-bit markers, managed at runtime using a \texttt{tilted} curation policy, which favors dense retention of more recent marker data \citep{moreno2024structured}.

To generate data representative of differing evolutionary conditions, two treatments were applied: an
\textit{adaptive regime} and a \textit{purifying regime}.
Under the adaptive regime, mutations increasing fitness were allowed, introducing the possibility of selective sweeps.
By contrast, the purifying regime treatment allowed only deleterious mutations.
Previous work with this system demonstrated the purifying regime to exhibit substantially greater phylogenetic richness, in line with expectations for selective sweeps to diminish population-level phylogenetic diversity \citep{moreno2024trackable}.

The simulation was run for 5 million generation cycles.
A per-PE population size of 256 organisms was used, providing a net population size of 190.8 million organisms.
``Fossil'' genome data was sampled on a rolling basis over the course of simulation using asynchronous \texttt{memcopy} operations, where sample genomes were copied from each PE on a rolling basis.
In total, collection of 1,999 (purifying regime) and 2,014 (adaptive regime) \texttt{memcopy} cycles elapsed.
Real-time runtime under each treatment was 24 minutes.

For implementation-level details on the Wafer-Scale Engine simulation, see \citep{moreno2024trackable}.

\subsection{Microbenchmark Experiments}

In a first set of microbenchmark trials, we performed a head-to-head comparison of runtime under the proposed shortcut-enabled algorithm versus the previous naive approach.
Given performance limitations of the naive approach, the maximum reconstruction workload tested was limited to 10,000 genomes.
Experiments used random subsamples of genomes from WSE fossil data described above.
To facilitate informative comparison, timings measured just the trie-building component of reconstructions, excluding pre- and post-processing steps.
Five replicate observations were taken per timing.

In a second set of microbenchmark trials, we conducted an empirical scaling analysis of the shortcut-enabled trie building algorithm across a wider spread of reconstruction workload sizes.
To assess the relationship between marginal throughput and trie size, we performed trie reconstructions of 10 million WSE genomes, broken into successive batches of 100,000 genomes.
This approach allowed us to independently measure time taken to insert each batch, to determine how this time changes for later batches (i.e., larger trees).
We also performed unbatched tests evaluating net execution time to build tries ranging from 1 to 10 million tips.
For these experiments, 3 replicates were performed per observation.

All microbenchmarks were replicated using genomes generated under adaptive and purifying regimes to assess the sensitivity of performance to underlying phylogeny structure.

\subsection{Macrobenchmark Experiments}

To assess the performance of our approach on large-scale workloads under real-world conditions, we supplemented microbenchmarks focusing on core trie extension and consolidation operations with two macrobenchmark trials comprising our full end-to-end reconstruction implementation (detailed below).
To reduce peak memory use, we additionally included a step to collapse unnecessary unifurcations (i.e., parent nodes with only a single child) between trie-building batches --- for more details, see the supplemental \citep{supplemental}.
These end-to-end experiments used the \texttt{hstrat.dataframe.surface\_build\_tree} CLI module incorporated in the \texttt{hstrat} library.

These experiments were conducted using the full billion-genome corpus of fossil data gathered from Wafer-Scale Engine simulations.
We performed two trial reconstructions: one using genome data from the adaptive regime treatment and another under the purifying regime treatment.
Reconstructions were conducted on 100-core cluster node allocations with 2 TB of memory.

\subsection{Validation Tests}

By design, our proposed shortcut-enabled algorithm produces equivalent reconstruction results compared to the reference naive approach, up to arbitrary tie-breaking decisions (see supplemental material \citep{supplemental} for detailed discussion).
To confirm correctness of our implementation, however, we report several validation tests.%
\footnote{In addition to validation experiments reported here, the \texttt{hstrat} library includes an extensive unit tests that compare reconstructed most recent common ancestor (MRCA) between organisms to the value expected by direct comparison of their hstrat marker records.}

Validation tests employed small evolutionary simulations run locally, to allow collection of ground-truth data via \texttt{phylotrackpy} \citep{dolson2024phylotrack}.
Simulations propagated populations of 100 agents under neutral evolution conditions, with runs lasting 500 generations.
We configured these simulations to consider both synchronous and asynchronous generations.
Reconstructions over end-state genomes were tested, as well as reconstructions incorporating fossil genomes sampled at 50-generation intervals.
To assess reconstruction quality, we calculated error as triplet distance between the ground truth and reconstructed phylogenies \citep{critchlow1996triples}.

\subsection{End-to-end Reconstruction Implementation}
\label{sec:pipeline}

\begin{figure*}
\centering
\begin{minipage}{0.7\linewidth}
\includegraphics[width=\linewidth]{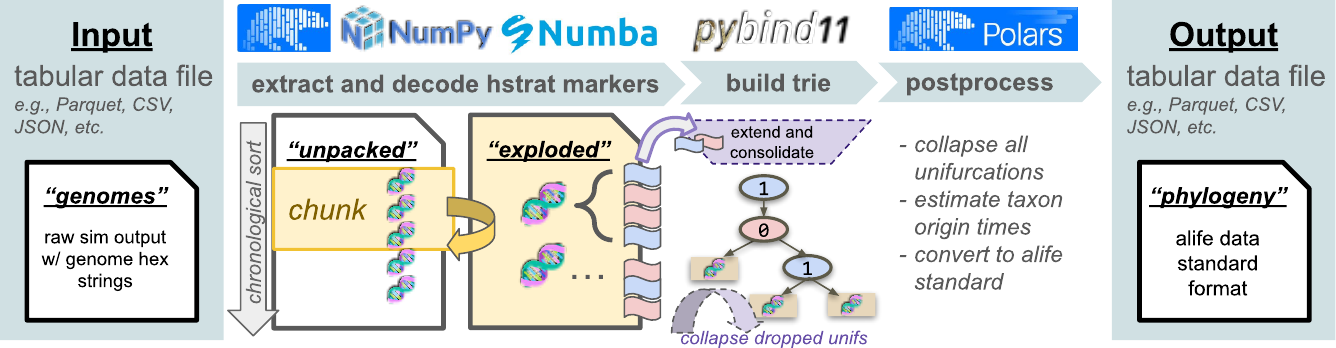}
\end{minipage}
\begin{minipage}{0.1\linewidth}\end{minipage}
\begin{minipage}{0.1\linewidth}\end{minipage}
\begin{minipage}{0.28\linewidth}
\vspace{1em}
\caption{
\textbf{Phylogeny pipeline.}
\small
Hstrat markers are ``unpacked'' from genome hex strings.
Individual markers are decoded into ``exploded'' format.
Inferred trie are progressively extended with genomes, purging dropped unifurcations.
Finally, trie is exported to ALife data standard.
High-performance scientific computing libraries are leveraged to support large datasets.
}
\label{fig:hstratpipeline}
\end{minipage}
\vspace{-.5em}
\end{figure*}

Figure \ref{fig:hstratpipeline} overviews the full end-to-end pipeline used to create a phylogenetic reconstruction from raw hstrat-annotated genome data.
First, hstrat annotation data is extracted and decoded.
To take advantage of SIMD parallelism, the decoding of origin times for genome markers is performed using bulk operations orchestrated via NumPy, batched over available processors using the Numba library threading engine.
Due to memory intensity of the exploded representation, where each genome marker constitutes an individual dataframe row, a chunk size may be configured to limit the amount of data exploded at any one time.
A chunk size of 50 million genomes was used for macrobenchmark trials.

Subsequently, decoded marker data is fed genome-by-genome into the trie-building backend.
To ensure full hardware utilization, execution of the core trie-building procedure --- which is single-threaded --- is performed concurrently with decode/explode work on the upcoming chunk.
Finally, a phylogeny is extracted from the end-state trie, converted to ALife standard format, and saved to disk.
The Polars library is used for load/save and other pre-/post-processing operations, allowing some operations to be distributed over available processor cores by the underlying threading engine.

\subsection{Software, Data, and Materials} \label{sec:materials}

Software used in this work is available via GitHub and archived via Zenodo.%
\footnote{\scriptsize Benchmarking and validation code (v1.0.1) is available on GitHub at \href{https://github.com/mmore500/hstrat-reconstruction-algo/tree/v1.0.1}{\texttt{mmore500/hstrat-reconstruction-algo}} \citep{matthew_andres_moreno_2025_16898918}, WSE kernel code (v2025.12.26) for Cerebras SDK v1.0.0 at \href{https://github.com/mmore500/wse-async-ga/tree/v2025.12.26}{\texttt{mmore500/wse-async-ga}} \citep{moreno_2025_16898904}, and hstrat library code (v1.20.10) at \href{https://github.com/mmore500/hstrat/tree/v1.20.10}{\texttt{mmore500/hstrat}} \citep{moreno_2025_16898849}.}
Data and supplementary material are hosted via the Open Science Framework at \href{https://osf.io/63ucz}{\texttt{osf.io/63ucz}} \citep{supplemental,foster2017open}.
All accompanying materials are provided open-source under the MIT License.

Microbenchmark experiments were conducted on a 2019 MacBook Pro (2.3GHz 8-Core Intel i9, 16GB RAM) using a benchmark script provided in supplemental material \citep{supplemental}.
Head-to-head trie-building comparisons were performed using Python implementations of both algorithms.
Empirical scaling analysis experiments used a C++ implementation of the shortcut-enabled trie-building algorithm.

Macrobenchmarks were performed on AMD EPYC compute cluster nodes with 128 cores and 2005 GB of memory.
Purifying- and adaptive-regime reconstructions were performed on separate nodes with EPYC 7H12 (2.595 GHz) and EPYC 7763 (2.445 GHz) processors, respectively.
End-to-end experiments also used a C++ implementation of the shortcut-enabled trie-building algorithm.

This project benefited significantly from open-source scientific software \citep{2020SciPy-NMeth,harris2020array,reback2020pandas,mckinney2010data,waskom2021seaborn,hunter2007matplotlib,moreno2022hstrat,yang2025downstream}.

\section{Results and Discussion} \label{sec:results}

\subsection{Microbenchmark: Runtime Comparison}

\begin{figure}[h]
\centering
\includegraphics[width=0.93\linewidth]{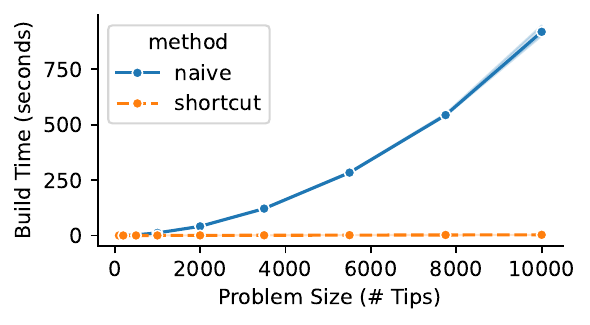}
\vspace{-1em}
\caption{
\textbf{Microbenchmark comparison of naive trie and shortcut table algorithms.}
\small
Naive approach appears to exhibit superlinear scaling.
Shaded bands are 100\% percentile intervals, with 5 samples per observation.
}
\label{fig:comparison}
\end{figure}

Direct comparison of runtimes across a range of problem sizes demonstrated substantial improvements under shortcut-enabled trie building, compared to the naive approach.
Figure~\ref{fig:comparison} compares execution time for both approaches for problem sizes ranging up to 10,000 genomes.
Over this window, the naive algorithm's runtime appears to grow at a superlinear pace.
At the largest surveyed problem size, naive runtime reaches 15 minutes.
In contrast, runtime for the shortcut-enabled approach at this scale is approximately 3 seconds --- a 300-fold difference in performance.

Although the same programming language was used for both implementations in this trial to ensure apples-to-apples comparability, implementations did differ in the underlying data structure used to store trie data.
The shortcut-enabled implementation used a contiguous edge-list table to store trie data, while the naive implementation used a node-and-pointer approach.
Typically, table-based approaches are faster in practice due to cache locality effects and reduced memory allocation calls.
However, both data structures provide equivalent time complexities for all relevant initialization and lookup operations.
Thus, the notable difference in scaling properties between the shortcut-enabled and naive benchmarks is indicative of meaningful differences in the underlying algorithms, beyond effects attributable to data structure implementations.

\subsection{Microbenchmark: Empirical Scaling Analysis}

\begin{figure*}[t]
  \centering
  \begin{minipage}{0.33\textwidth}
    \centering
    \includegraphics[height=4.3cm]{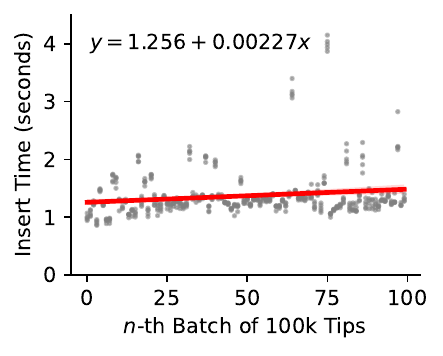}
    \vspace{-0.5em}
    \subcaption{Purifying Regime}
    \label{fig:scaling:purifying}
  \end{minipage}%
  \begin{minipage}{0.33\textwidth}
    \centering
    \includegraphics[height=4.3cm,trim={0.7cm 0 0 0},clip]{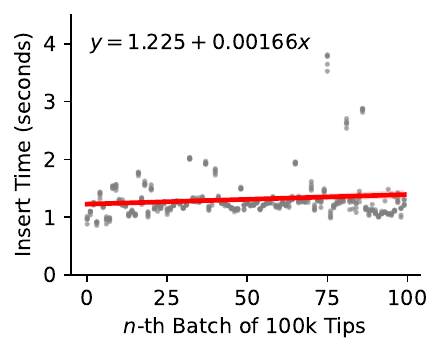}
    \vspace{-0.5em}
    \subcaption{Adaptive Regime}
    \label{fig:scaling:adaptive}
  \end{minipage}%
  \begin{minipage}{0.33\textwidth}
    \caption{%
    \textbf{Empirical performance scaling of shortcut table algorithm in microbenchmark trials.} \small
    Panels \ref{fig:scaling:purifying} and \ref{fig:scaling:adaptive} differ in simulation data source of sampled genomes, with the former exhibiting higher phylogenetic richness. Data collected as batches are added in a 10 million tip reconstruction, sampled 5 times per panel. \vspace{2.5em}
}
    \label{fig:scaling}
  \end{minipage}
  \vspace{-1.5em}
\end{figure*}

Having seen a substantial speedup in comparing the shortcut-enabled algorithm to the naive trie-building approach, we next sought to assess the scaling properties of the shortcut-enabled approach in isolation.

Figure~\ref{fig:scaling} depicts the relationship between trie size as 100,000-tip batches are added and the time taken to insert each batch, over 5 replicates of 10 million tip reconstructions.
Fitting a linear regression model to this data, we found significant evidence to reject the null hypothesis of a constant slope, for both the adaptive and purifying regime datasets ($\alpha = 0.05$).
Evidence indicates, therefore, that as trie size increases, the time required per tip added increases as well.

Notably, batch processing times exhibit strong outliers for which trie extension was notably slower.
This effect is due to the presence of expensive consolidation steps within those batches.
To assess this phenomenon, we performed a more fine-grained analysis (with batches of 5,000 tips) on non-outlier data (filtering out values above the 95th percentile).
Further regression analysis using this data suggests that, past a certain point, non-outlier batch times do not increase significantly with larger tree size.
(See supplement for more information \citep{supplemental}.)
In an analytical investigation of scaling complexity, which is left to future work, we therefore expect the degree to which expensive trie consolidation operations are amortized over average-case insertions to play an important role in determining the scaling complexity of the shortcut-enabled approach.

Microbenchmarks measuring net reconstruction time across problem sizes, rather than marginal batch-wise throughput, reach a similar conclusion and are provided in Supplemental Figure~\ref{fig:asymptotic} \citep{supplemental}.

\subsection{Macrobenchmark: Billion-Tip Reconstruction}

\begin{figure}[h]
\centering
\includegraphics[width=0.8\linewidth]{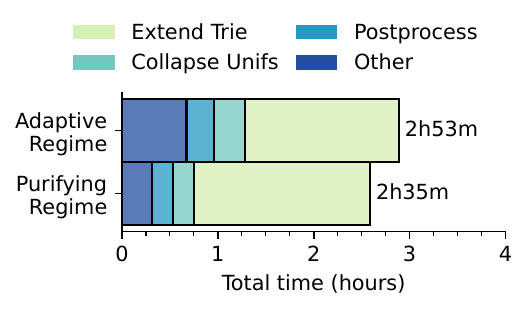}
\vspace{-1em}
\caption{%
\textbf{Execution profile of reconstruction pipeline from macrobenchmark trial.}
\small
Results are shown for 1 billion tip reconstruction trials.
Trials differed in simulation data source, with ``adaptive regime'' genomes sampled under conditions with lower phylogenetic richness.
Other category includes file load and save time as well as hstrat marker explosion/decoding.
}
\label{fig:billion-tip-time}
\end{figure}

Given the promising results of microbenchmark trials, particularly the favorable empirical scaling result, we next sought to assess the performance of our approach on an extremely large problem size.

For this purpose, we conducted trial reconstruction benchmarks comprising the full corpus of 1 billion agent genomes extracted from our trial Wafer-Scale Engine experiments.
As a point of comparison, the largest reconstructions performed using the previous approach comprised only 10,000 tips \citep{moreno2024trackable}.

Benchmarked operations included the full reconstruction operation pipeline shown in Figure \ref{fig:hstratschematic}.
Input data comprised a Parquet format file storing raw agent genomes as hexadecimal strings.
The population file size was 34 GB for and 35 GB for the sample adaptive- and purifying-selection regimes, respectively.
The output data comprised the reconstructed phylogeny in an edge-list format closely resembling the ALife data standard.
Several pieces of metadata (e.g., agent position) were forwarded to the output data.
For the adaptive-selection regime, the output file size was 54 GB, and for the purifying-selection regime, the output file size was 55 GB.

Given the large problem-size scale tested, macrobenchmark trials were carried out on a compute cluster node, rather than a desktop PC, as was used for microbenchmark experiments.
Figure \ref{fig:billion-tip-time} profiles net runtime in our case study reconstruction trials.
Reconstruction time was 2h:53m for the adaptive-regime data set and 2h:35m for the purifying-regime data set.
These figures correspond to a net throughput of 5.8 and 6.4 million tips per minute, respectively.
Peak memory use was 693 GB (adaptive regime) and 614 GB (purifying regime).

\subsection{Validation: Comparison to Ground Truth}

In a first set of trials to validate the quality of phylogenies produced by the shortcut-enabled algorithm, we tested inference accuracy against ground-truth data.
For these trials, we conducted experiments using a neutral evolution model on a single CPU, where we were able to directly record the underlying phylogeny history.
Example visualizations of corresponding reconstruction and ground-truth phylogenies are provided in supplementary materials \citep{supplemental}.

\begin{table}[h]
\centering
\begin{tabular}{r|c|c|c|c}
Surface Size & 8 & 16 & 32 & 64  \\
\hline 
Mean Error & 0.373 & 0.191 & 0.095 & 0.021 \\
Standard Deviation & 0.240 & 0.156 & 0.082 & 0.018 \vspace{-1ex}
\end{tabular}
\caption{\textbf{Accuracy of shortcut table algorithm on simulated data.}
\small Mean error is defined as the average triplet distance between reconstructed and ground truth phylogenies over 20 random simulations of evolution, with each organism having a variable surface size. A steady retention policy was used.
}
\label{table:validation}

\end{table}

Table~\ref{table:validation} reports reconstruction error across annotation sizes ranging from 8 single-bit markers up to 64 single-bit markers.
As expected, medium and large annotation sizes produced lowest reconstruction error, with large annotations of 64 bits achieving triplet distance error on average around 2\%.%
\footnote{This error measure corresponds to the fraction of sampled three-node sets for which both trees report the same two nodes as more closely related.}
In contrast, very small 8-bit annotations averaged 37\% reconstruction error.

\section{Conclusion} \label{sec:conclusion}

We have presented and benchmarked a new algorithm for phylogenetic reconstruction from synthetic hstrat \citep{moreno2024hstrat} data with significantly better performance, both empirically and asymptotically, rectifying the existing bottleneck of slow phylogenetic reconstruction in wafer-scale distributed simulations.

The scope of scientific questions tractable to investigation \textit{in silico} continues to be broadened by rapid advances in computing technology.
However, such advances also pose new challenges in managing vast quantities of data.
By providing scalable means for fast phyloanalysis of large simulations, our work seeks to help the scientific utility of parallel and distributed digital evolution experiments keep pace with opportunities afforded by emerging hardware architectures.

In a parallel vein, the volume of data processed in bioinformatics workflows is also increasing with continuing advances in high-throughput sequencing technologies, enabling the construction of phylogenies containing millions of taxa.
As an illustrative example at the cutting edge of extreme scale, \citet{konno2022deep} reports phylogeny synthesis from 235 million sequence reads generated from an \textit{in silico} CRISPR barcoding model --- requiring 31 hours of compute time across 300 HPC nodes.

In both the context of bioinformatics and artificial life research, very large-scale phylogeny data enabled by advances in sequencing and computing technology represent a new challenge as much as an opportunity, raising the question of how best to mine this data.
On a practical level, work is needed not just to push the boundaries of what can be learned from phylogenies, but also how to store, load, traverse, quantify, visualize, and manipulate very large phylogenies in an efficient manner.
Indeed, projects are being developed to try to address this issue.
For example, taxonium \citep{sanderson2022taxonium} is a web-based software for visualizing large phylogenies in a flexible, interactive manner, and is able to handle browsing millions of tips at a high frame rate.
Other projects aim to create methods for compact, scalable phylogeny representations \citep{moshiri2025compacttree, moshiri2020treeswift}, enabling faster and more memory-efficient tree operations.

\subsection{Future Work}

In pushing the boundaries of phylogenetic scale to billion-tip datasets, ALife research has the opportunity to contribute to an interdisciplinary ecosystem of software tools developing around working with very large-scale phylogenies.
In particular, the ALife data standard, which specifies a tabular representation for phylogeny data \citep{Lalejini2019data}, has strong potential to contribute to a larger high-performance phylogeny processing infrastructure.
Although originally envisioned as a data storage format, the tabular nature of the standard allows integrations with high-performance software tools built around the ``dataframe'' concept, including Pandas, Polars, Dask, and data.table.
These libraries provide a structured, user-friendly interface to advanced performance features such as multithreading, data streaming, query optimization, file partitioning, and column-oriented binary file formats \citep{mckinney2010data,datatable,vink2024polars,rocklin2015dask}.
Additionally, for Python users, the columnar array format typical in dataframe libraries is compatible with NumPy and Numba, readily enabling on-the-fly SIMD vectorization and just-in-time compilation \citep{harris2020array,lam2015numba}.
Indeed, this approach underlies much of the pre- and post-processing steps for end-to-end reconstruction demonstrated in this work.

Of more direct bearing to phylogeny reconstruction from heredity stratigraphy data, future work should also assess trade-offs in configuring hstrat annotations for phylogeny reconstructions incorporating fossil data (including extinct lineages) drawn from earlier time points, as was the case in this work.
Whereas existing analysis has focused on reconstructions from genomes drawn from a shared contemporary population \citep{moreno2025testing}, it is likely that configuration adjustments may be necessary for scenarios involving genomes of widely varying generational depths.
Promisingly, in preliminary testing, we have found indications that incorporating fossil data into reconstructions can help improve inference accuracy of relationships among extant population members.

Other future work will involve experiments putting the high-throughput phylogeny reconstruction capabilities developed into practice.
In ongoing work, we are interested in using large-scale digital evolution experiments to test phylostatistical methodologies for detecting signatures of multilevel selection associated with major transitions in evolution \citep{BonettiFranceschi2024}, as well as pursuing hypothesis-driven experiments investigating mechanisms of open-ended evolution.
To this end, all tools described herein are published as modular open source library components, supporting the broader research community in exploring and extending this line of inquiry.

\section*{Acknowledgment}
\begin{footnotesize}
Computational resources were provided by the MSU Institute for Cyber-Enabled Research and by the Pittsburgh Supercomputing Center Neocortex under NSF ACCESS Innovative Projects Allocation BIO240102 \citep{buitrago2021neocortex,boerner2023pearc}.
Any opinions, findings, and conclusions or recommendations expressed in this material are those of the author(s) and do not necessarily reflect the views of the National Science Foundation.

This material is based upon work supported by the U.S. Department of Energy, Office of Science, Office of Advanced Scientific Computing Research (ASCR), under Award Number DE-SC0025634.
This report was prepared as an account of work sponsored by an agency of the United States Government.
Neither the United States Government nor any agency thereof, nor any of their employees, makes any warranty, express or implied, or assumes any legal liability or responsibility for the accuracy, completeness, or usefulness of any information, apparatus, product, or process disclosed, or represents that its use would not infringe privately owned rights.
Reference herein to any specific commercial product, process, or service by trade name, trademark, manufacturer, or otherwise does not necessarily constitute or imply its endorsement, recommendation, or favoring by the United States Government or any agency thereof.
The views and opinions of authors expressed herein do not necessarily state or reflect those of the United States Government or any agency thereof.

This material is based upon work supported by the Eric and Wendy Schmidt AI in Science Postdoctoral Fellowship, a Schmidt Sciences program.
\end{footnotesize}

\putbib

\end{bibunit}

\clearpage
\newpage

\begin{bibunit}

\appendix
\section*{\Huge Supplemental Material for ``A Scalable Trie Building Algorithm for High-Throughput Phyloanalysis of Wafer-Scale Digital Evolution Experiments''}

\setcounter{section}{0}

\makeatletter
\def\@seccntformat#1{\@ifundefined{#1@cntformat}%
   {\csname the#1\endcsname\space}
   {\csname #1@cntformat\endcsname}}
\newcommand\section@cntformat{\thesection.\space} 
\makeatother
\renewcommand{\thesection}{S\arabic{section}}
\counterwithin{equation}{section}
\counterwithin{figure}{section}
\counterwithin{table}{section}
\counterwithin{theorem}{section}
\counterwithin{algorithm}{section}
\counterwithin{lstlisting}{section}

\vspace{0.8em}

Below is supplemental material for our paper \citep{supplemental} with discussions and figures that were not put in the main paper.

\subsection{Naive Algorithm Definition}

\label{sec:algorithm:naivedefinition}

As stated in the original work, we use the concept of the ``longest successive streak'' \citep{moreno2024analysis} to determine the most likely path to follow in the trie when adding some organism.
We do this by performing the \textsc{MostLikelyChild} step as described in Algorithm~\ref{alg:old} in the main paper, defined here in Algorithm~\ref{alg:mostlikely}.

\begin{algorithm}[h]
    \begin{algorithmic}[1]
    \small{
        \Function{DifferentiaAtRank}{organism $o$, rank $r$}
            \For{$(r', d') \in o$}
                \If{$r' = r$}
                    \Return $d'$
                \EndIf
            \EndFor
            \State \Return $-1$
        \EndFunction
        \Function{DeepestRank}{node $n$, organism $o$, rank $r$}
            \State $d \gets \textsc{DifferentiaAtRank}(o, \operatorname{rank}(n))$
            \If{$d \ne -1 \land d \ne \operatorname{differentia}(n)$}
                \State \Return $r$
            \EndIf
            \If{$|\operatorname{children}(n)| = 0$}
                \State \Return $r$
            \EndIf
            \State \Return $\max_{c \in \operatorname{children}(n)} \textsc{DeepestRank}(c, o, \operatorname{rank}(n))$
        \EndFunction
        \Function{MostLikelyChild}{node $n$, organism $o$}
            \State $c' \gets \arg\max_{c \in \operatorname{children}(n)} \textsc{DeepestRank}(c, o, \operatorname{rank}(n))$
            \If{$\textsc{DeepestRank}(c', o, \operatorname{rank}(n)) = \operatorname{rank}(n)$}
                \State \Return $n$
            \EndIf
            \State \Return $c'$
        \EndFunction
    }
    \end{algorithmic}
    \caption{\textbf{The most likely child algorithm.} \small Determines the child of $n$ most likely to contain the organism $o$ somewhere in its subtree, returning $n$ itself if there are no candidate children. \vspace{-1.5em}}
    \label{alg:mostlikely}
\end{algorithm}

\subsection{Algorithm Equivalency}

\label{sec:algorithm:equivalency}

We shall demonstrate the accuracy of the proposed shortcut algorithm by providing a proof overview of how it is equivalent to the old algorithm (and assuming the old algorithm is a `good' reconstruction algorithm).
We define `equivalent' in this context to mean that every node at which an organism can be placed by the naive algorithm is also a node at which one can be placed by the shortcut algorithm (i.e. the shortcut algorithm places nodes at points that the naive algorithm could, given the right arbitrary choices).

Consider adding some organism $o$ to the reconstructed tree $T$, which follows exactly one of the following cases: there is no missing information (i.e. $r = \operatorname{rank}(\operatorname{children}(n) \forall (r, d) \in o)$, where $n$ is the current node throughout iteration), or there is.
The case where there is no missing information is trivial -- the only difference between the original and new algorithm lies when there is missing information, the absence of which rendering them the same.

Therefore, consider analyzing data at rank $r$ and having some missing information between the $\operatorname{rank}(n)$ and $r$.
Now, consider cases on whether there is some successive path of matching differentiae:

\begin{itemize}
  \item \textbf{There is some path.}
  Therefore, we know that this path contains a descendant $n'$ of $n$ with rank $r$ and differentia $d$.
  In the naive algorithm, this descendant is selected to be the next $n$ when processing the information of $o$, as the optimal path found includes $n'$.
  However, we also know that the shortcut algorithm must have created a shortcut from $n$ to $n'$ (and possibly collapsed it with other valid descendants, which is irrelevant as the path still remains).
  So, the shortcut algorithm (in line 10 of the naive Algorithm~\ref{alg:old} as described in the paper) finds this child and selects it to be the new $n$ as well.
  \item \textbf{There is no such path.}
  Since there is no candidate path to follow, the naive algorithm simply creates a new node branching off $n$ with rank and differentia $(r, d)$.
  Then, this new node is made the new $n$.
  Since there was no valid path, we see that $n$ has no descendants with rank $r$ and differentia $d$.
  Therefore, there is no shortcut built from $n$ to such a node, as it does not exist.
  So, the shortcut algorithm will also branch off of $n$ by creating a new node in the same manner as the naive one.
\end{itemize}

In either case, we have shown that both algorithms make effectively the same decisions when it comes to selecting nodes $n$ throughout each iteration of \textsc{TreeInsert}.
Because original relationships between nodes are still preserved, choosing a particular node to branch off of conveys identical information in both algorithms.
In conclusion, the two algorithms are equivalent in correctness.
However, it is worth noting that in cases where multiple paths are valid, the algorithms may make different decisions (as the chosen path out of these is arbitrary) and therefore may not be exactly equivalent.

\subsection{Algorithm Implementations}

The shortcut-based algorithm is implemented in the \texttt{hstrat} Python package \citep{moreno2022hstrat} as \texttt{hstrat.build\_tree\_searchtable}.
The reconstruction backend is selectable between independent implementations written in either pure Python or in C++, bound via Pybind11 \citep{wenzel2017pybind11}.
Benchmarks comparing the shortcut-enabled approaches to the existing naive approach, implemented in pure Python, used the pure Python backend for the shortcut-enabled approach, to ensure apples-to-apples comparability.
Other benchmarking and validation experiments used the C++ backend, which is expected to be better reflective of how the library will be used in practice, given the inherent speedup and memory savings associated with compiled code.

As suggested by the function name, both shortcut-enabled backends represent trie data using an edge-list table, rather than a node-and-pointer approach, as is the case with the existing naive implementation.
From an algorithmic perspective, tabular and node-and-pointer data structures provide identical properties for traversal and lookup operations.
However, in practice, the former approach is often faster due to fewer memory allocation operations and better cache locality.

At the implementation level, we represent trie shortcuts via additional columns in our edge table.
When a new node is appended, its shortcut edge is initialized identically to its trie edge.
From this point onwards, shortcut edges are used only for traversing the trie, while underlying trie edges --- treated as immutable from node initialization onwards --- are used when reading out the final reconstruction result.

\subsection{Build-time Garbage Collection}

To reduce memory usage for very large phylogeny reconstructions, we supplement the core trie extension and consolidation procedure detailed in Section~\ref{sec:algorithm} with an intermittent garbage collection step.
Because genomes are inserted in ascending order of generational depth, trie nodes at a generation with a single child node.
Given that all trie nodes only have a single parent node, such a ``dropped unifurcation node'' can be excised and safely replaced with an edge directly connecting its parent and child.

As currently implemented, we use the presence of modifications to shortcut edges to identify nodes for which corresponding hstrat markers have been dropped.
We note that this does not detect all dropped unifurcations, and an alternate more aggressive approach could be implemented readily by incorporating information as to which markers have been dropped from the marker retention algorithm used at runtime.

At implementation-level, the frequency of ``dropped unifurcation'' garbage collection is provided as a configurable function parameter, and so can be tuned as appropriate for a given use case scenario.
Garbage collection was not necessary for the up to 10 million tip reconstructions evaluated in our microbenchmarking experiment, and so was not performed.
For our macrobenchmark experiments, we ran this procedure at three evenly-spaced intervals during the build process.
Figure~\ref{fig:billion-tip-time} details the share of net execution time consumed by purging dropped unifurcations in these trials.

\subsection{Validation with Ground Truth}

In addition to determining triplet distance between ground truth and reconstructed phylogenies, we ran visualizations to make comparisons more interpretable.

\begin{figure}[h]
\includegraphics[width=\columnwidth,trim={0 52cm 0 0.8cm},clip]{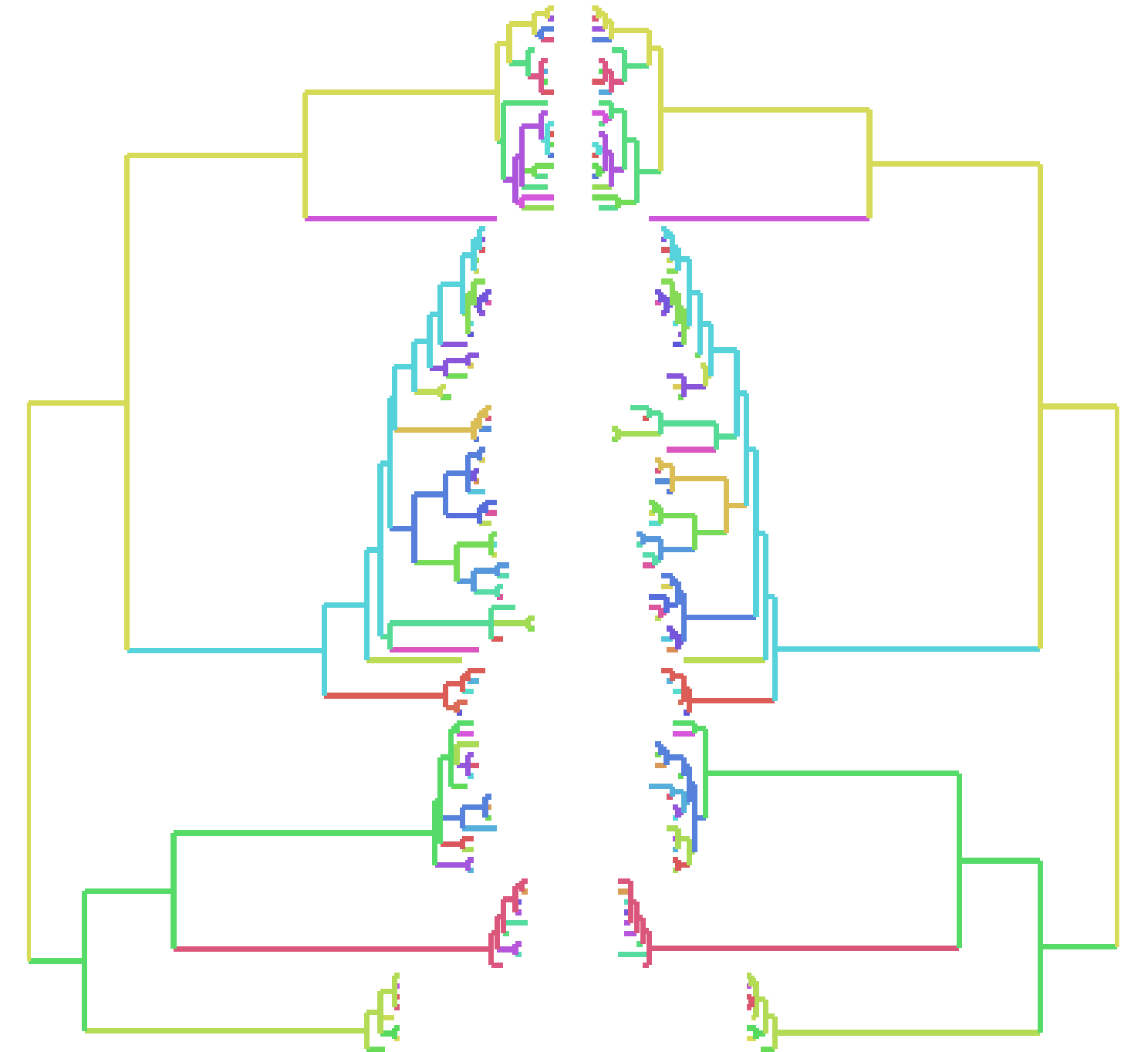}
\caption{\textbf{Sample comparison of true and reconstructed phylogenies.}
\small
Generated using a tilted retention policy and a surface of 32 bits. The true phylogeny is on the left and the reconstructed phylogeny is on the right. Colors are based on a hash from the taxon label for each tip to better facilitate visual comparison. Reconstruction error for this reconstruction was 2.6\%. Visualization created with colorclade \citep{moreno2024colorclade}.
}
\label{fig:colorclade}
\end{figure}

\subsection{More on Microbenchmark Scaling}

We found a positive linear relationship between the speed at which tips were processed with the number of tips already present (i.e. the position of the batch).
However, this was when large outliers are taken into account; by removing them, we see a much more nuanced relationship.
The average-case insertions seem to be increasing up to a certain point, after which they, by and large, taper off in runtime (this pattern is likely due to hardware details enabling faster additions for small trees).
In fact, we found that there was a statistically significant breakpoint through piecewise regression \citep{pilgrim2021piecewise,davies1987hypothesis}.

Given this, we decided on a hardcoded breakpoint of 3 million tips --- that is, when applying piecewise linear regression, we would consider two different lines for the cases where we are under or over 3 million tips.
This hardcoded breakpoint was chosen (as opposed to that generated by the regression) in order to prevent overfitting on a certain model.
Regardless, we found that our chosen breakpoint is also transferable to other instances of reconstruction, such as with a purifying regime --- we see the same pattern of increasing insertion time that begins to taper off past about 3 million tips.

\begin{figure*}[h]
  \includegraphics[width=\textwidth]{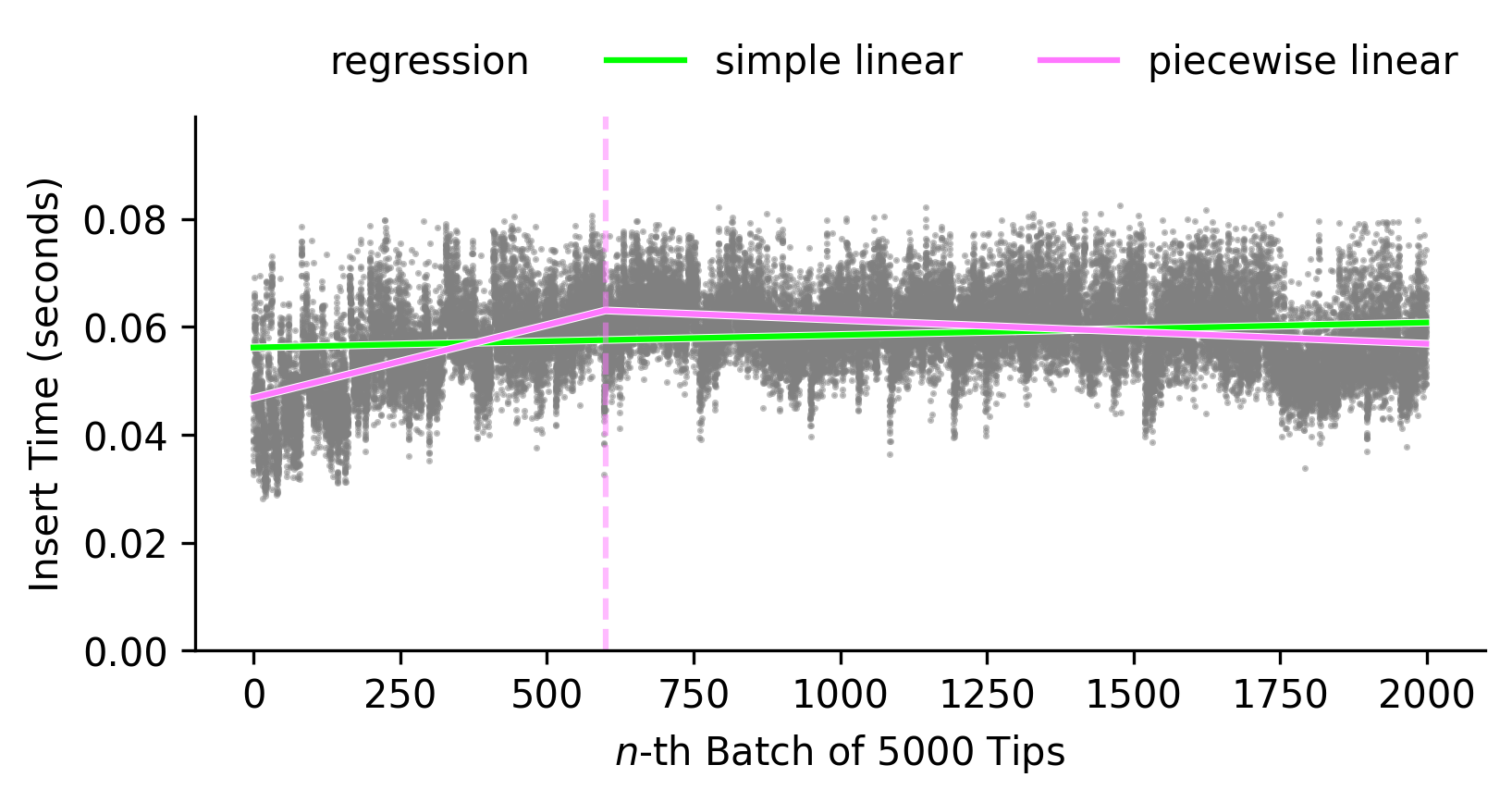}
  \caption{\textbf{Time taken per batch with outliers removed.} \small Graph generated by running the same 10 million tip reconstruction 20 times, timing the insertions of each batch of 5000 tips. Breakpoint in piecewise linear regression hardcoded at 3 million tips (i.e. the 600-th batch of 5000 tips).}
  \label{fig:average-scaling}
\end{figure*}

Given this, we then fitted a piecewise linear regression on the data, with a breakpoint at 600 (representing 3 million tips, as batches were of size 5000).
We found that we could not reject the null hypothesis that the the piece corresponding to entries added to a tree with more than 3 million tips had a nonzero slope -- in fact the linear regression modeled it as a negative slope.
For a comparison of simple linear regression with piecewise linear regression on the average case data, see Figure~\ref{fig:average-scaling}.

Regardless, this analysis is merely on the average, and it is very important to take into account the outliers during overall analysis, as they represent the bulk of the work done by the algorithm.

\begin{figure*}[h]
  \centering
  \begin{minipage}{0.33\textwidth}
    \centering
    \includegraphics[height=4.3cm]{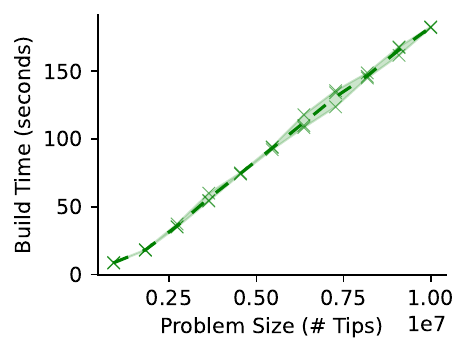}
    \subcaption{Purifying Regime}
    \label{fig:asymptotic:purifying}
  \end{minipage}%
  \begin{minipage}{0.33\textwidth}
    \centering
    \includegraphics[height=4.3cm,trim={0.7cm 0 0 0},clip]{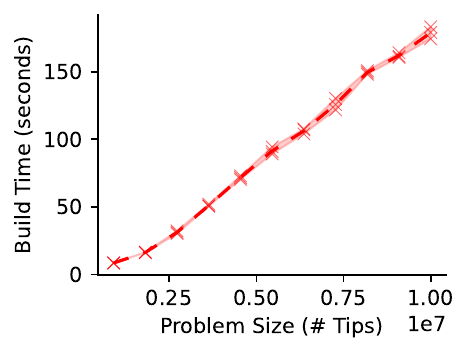}
    \subcaption{Adaptive Regime}
    \label{fig:asymptotic:adaptive}
  \end{minipage}%
  \begin{minipage}{0.33\textwidth}
    \caption{%
    \textbf{Empirical performance scaling of shortcut table algorithm in microbenchmark trials.} \small
    Panels \ref{fig:asymptotic:purifying} and \ref{fig:asymptotic:adaptive} differ in simulation data source of sampled genomes, with the former exhibiting higher phylogenetic richness. Shaded bands are 100\% percentile intervals, with 3 samples taken per problem size.\vspace{2.5em}
}
    \label{fig:asymptotic}
  \end{minipage}
\end{figure*}

When measuring the time taken on entire reconstructions with varying numbers of tips, we found an approximately linear relationship between the two variables as depicted in Figure~\ref{fig:asymptotic}. 
This seems to suggest that the outliers detected during the empirical analysis in the main paper are being amortized away by the distance between them. 
However, we can make no conclusive claim without a more thorough analysis.

\putbib

\end{bibunit}


\begin{thebibliography}{}

\bibitem[Ackley, 2016]{ackley2016indefinite}
Ackley, D. (2016).
\newblock Indefinite scalability for living computation.
\newblock {\em Proceedings of the AAAI Conference on Artificial Intelligence},
  30(1), 4142--4146.
\newblock \url{https://doi.org/10.1609/aaai.v30i1.9802}

\bibitem[Ackley \& Small, 2014]{ackley2014indefinitely}
Ackley, D. \& Small, T. (2014).
\newblock Indefinitely scalable computing = artificial life engineering.
\newblock {\em Artificial Life 14: Proceedings of the Fourteenth International
  Conference on the Synthesis and Simulation of Living Systems}, Alife,
  606–613.
\newblock \url{https://doi.org/10.1162/978-0-262-32621-6-ch098}

\bibitem[Ackley, 2023]{ackley2023robust}
Ackley, D.~H. (2023).
\newblock A robust programmable replicator for an indefinitely scalable
  machine.
\newblock {\em The 2023 Conference on Artificial Life}, Alife 2023.
\newblock \url{https://doi.org/10.1162/isal_a_00701}

\bibitem[Barrett et~al., 2025]{datatable}
Barrett, T., Dowle, M., Srinivasan, A., Gorecki, J., Chirico, M., Hocking, T.,
  Schwendinger, B., \& Krylov, I. (2025).
\newblock {\em data.table: Extension of `data.frame`}.
\newblock \url{https://r-datatable.com}.
\newblock R package version 1.17.99, https://Rdatatable.gitlab.io/data.table,
  https://github.com/Rdatatable/data.table

\bibitem[Boerner et~al., 2023]{boerner2023pearc}
Boerner, T.~J., Deems, S., Furlani, T.~R., Knuth, S.~L., \& Towns, J. (2023).
\newblock Access: Advancing innovation: Nsf’s advanced cyberinfrastructure
  coordination ecosystem: Services \& support.
\newblock {\em Practice and Experience in Advanced Research Computing}, Pearc
  ’23, 173–176.
\newblock \url{https://doi.org/10.1145/3569951.3597559}

\bibitem[Bonetti~Franceschi \& Volz, 2024]{BonettiFranceschi2024}
Bonetti~Franceschi, V. \& Volz, E. (2024).
\newblock Phylogenetic signatures reveal multilevel selection and fitness costs
  in sars-cov-2.
\newblock {\em Wellcome Open Research}, 9, 85.
\newblock \url{https://doi.org/10.12688/wellcomeopenres.20704.2}

\bibitem[Buitrago \& Nystrom, 2021]{buitrago2021neocortex}
Buitrago, P.~A. \& Nystrom, N.~A. (2021).
\newblock {\em Neocortex and Bridges-2: A High Performance AI+HPC Ecosystem for
  Science, Discovery, and Societal Good}, 205–219.
\newblock Springer International Publishing.
\newblock \url{https://doi.org/10.1007/978-3-030-68035-0_15}

\bibitem[Colijn \& Gardy, 2014]{colijn2014phylogenetic}
Colijn, C. \& Gardy, J. (2014).
\newblock Phylogenetic tree shapes resolve disease transmission patterns.
\newblock {\em Evolution, medicine, and public health}, 2014(1), 96--108.

\bibitem[Critchlow et~al., 1996]{critchlow1996triples}
Critchlow, D.~E., Pearl, D.~K., \& Qian, C. (1996).
\newblock The triples distance for rooted bifurcating phylogenetic trees.
\newblock {\em Systematic Biology}, 45(3), 323--334.

\bibitem[Daudey et~al., 2024]{daudey2024aevol}
Daudey, H., Parsons, D.~P., Tannier, E., Daubin, V., Boussau, B., Liard, V.,
  Gallé, R., Rouzaud-Cornabas, J., \& Beslon, G. (2024).
\newblock Aevol\_4b: Bridging the gap between artificial life and
  bioinformatics.
\newblock {\em The 2024 Conference on Artificial Life}, Alife 2024.
\newblock \url{https://doi.org/10.1162/isal_a_00716}

\bibitem[De~La~Briandais, 1959]{delabriandais1959file}
De~La~Briandais, R. (1959).
\newblock File searching using variable length keys.
\newblock {\em Papers presented at the the March 3-5, 1959, western joint
  computer conference on XX - IRE-AIEE-ACM ’59 (Western)}, IRE-AIEE-ACM ’59
  (Western), 295–298.
\newblock \url{https://doi.org/10.1145/1457838.1457895}

\bibitem[Dolson \& Ofria, 2021]{dolson2021digital}
Dolson, E. \& Ofria, C. (2021).
\newblock Digital evolution for ecology research: A review.
\newblock {\em Frontiers in Ecology and Evolution}, 9.
\newblock \url{https://doi.org/10.3389/fevo.2021.750779}

\bibitem[Dolson et~al., 2024]{dolson2024phylotrack}
Dolson, E., Rodriguez-Papa, S., \& Moreno, M.~A. (2024).
\newblock {\em Phylotrack: C++ and python libraries for in silico phylogenetic
  tracking}.
\newblock \url{https://doi.org/10.48550/arxiv.2405.09389}

\bibitem[Emani et~al., 2024]{emani2024democratizing}
Emani, M., Raskar, S., John, J., Durillo, J.~J., Ben-Nun, T., Van~Essen, B.,
  Ltaief, H., \& Brown, N. (2024).
\newblock {\em Democratizing ai accelerators for hpc applications: Challenges,
  success, and support}.
\newblock
  \url{https://sc24.conference-program.com/presentation/?id=bof133&amp=&sess=sess628}

\bibitem[Foster \& Deardorff, 2017]{foster2017open}
Foster, E.~D. \& Deardorff, A. (2017).
\newblock Open science framework (osf).
\newblock {\em Journal of the Medical Library Association}, 105(2).
\newblock \url{https://doi.org/10.5195/jmla.2017.88}

\bibitem[Fredkin, 1960]{fredkin1960trie}
Fredkin, E. (1960).
\newblock Trie memory.
\newblock {\em Communications of the ACM}, 3(9), 490--499.
\newblock \url{https://doi.org/10.1145/367390.367400}

\bibitem[Genthon et~al., 2023]{genthon2023cell}
Genthon, A., Nozoe, T., Peliti, L., \& Lacoste, D. (2023).
\newblock Cell lineage statistics with incomplete population trees.
\newblock {\em PRX Life}, 1(1).
\newblock \url{https://doi.org/10.1103/prxlife.1.013014}

\bibitem[Haller \& Messer, 2023]{haller2023slim}
Haller, B.~C. \& Messer, P.~W. (2023).
\newblock Slim 4: Multispecies eco-evolutionary modeling.
\newblock {\em The American Naturalist}, 201(5), E127–e139.
\newblock \url{https://doi.org/10.1086/723601}

\bibitem[Harris et~al., 2020]{harris2020array}
Harris, C.~R., Millman, K.~J., van~der Walt, S.~J., Gommers, R., Virtanen, P.,
  Cournapeau, D., Wieser, E., Taylor, J., Berg, S., Smith, N.~J., Kern, R.,
  Picus, M., Hoyer, S., van Kerkwijk, M.~H., Brett, M., Haldane, A., del Río,
  J.~F., Wiebe, M., Peterson, P., Gérard-Marchant, P., Sheppard, K., Reddy,
  T., Weckesser, W., Abbasi, H., Gohlke, C., \& Oliphant, T.~E. (2020).
\newblock Array programming with numpy.
\newblock {\em Nature}, 585(7825), 357–362.
\newblock \url{https://doi.org/10.1038/s41586-020-2649-2}

\bibitem[Hernandez et~al., 2022]{hernandez2022phylogenetic}
Hernandez, J.~G., Lalejini, A., \& Dolson, E. (2022).
\newblock Phylogenetic diversity predicts future success in evolutionary
  computation.
\newblock {\em Proceedings of the Genetic and Evolutionary Computation
  Conference Companion}, Gecco ’22, 23–24.
\newblock \url{https://doi.org/10.1145/3520304.3534079}

\bibitem[Hunter, 2007]{hunter2007matplotlib}
Hunter, J.~D. (2007).
\newblock Matplotlib: A 2d graphics environment.
\newblock {\em Computing in Science \& Engineering}, 9(3), 90--95.
\newblock \url{https://doi.org/10.1109/mcse.2007.55}

\bibitem[Konno et~al., 2022]{konno2022deep}
Konno, N., Kijima, Y., Watano, K., Ishiguro, S., Ono, K., Tanaka, M., Mori, H.,
  Masuyama, N., Pratt, D., Ideker, T., Iwasaki, W., \& Yachie, N. (2022).
\newblock Deep distributed computing to reconstruct extremely large lineage
  trees.
\newblock {\em Nature Biotechnology}, 40(4), 566–575.
\newblock \url{https://doi.org/10.1038/s41587-021-01111-2}

\bibitem[Lalejini et~al., 2019]{Lalejini2019data}
Lalejini, A., Dolson, E., Bohm, C., Ferguson, A.~J., Parsons, D.~P., Rainford,
  P.~F., Richmond, P., \& Ofria, C. (2019).
\newblock Data standards for artificial life software.
\newblock {\em The 2019 Conference on Artificial Life}, Alife 2019.
\newblock \url{https://doi.org/10.1162/isal_a_00213}

\bibitem[Lam et~al., 2015]{lam2015numba}
Lam, S.~K., Pitrou, A., \& Seibert, S. (2015).
\newblock Numba: a llvm-based python jit compiler.
\newblock {\em Proceedings of the Second Workshop on the LLVM Compiler
  Infrastructure in HPC}, Sc15, 1–6.
\newblock \url{https://doi.org/10.1145/2833157.2833162}

\bibitem[Levy et~al., 2015]{levy2015quantitative}
Levy, S.~F., Blundell, J.~R., Venkataram, S., Petrov, D.~A., Fisher, D.~S., \&
  Sherlock, G. (2015).
\newblock Quantitative evolutionary dynamics using high-resolution lineage
  tracking.
\newblock {\em Nature}, 519(7542), 181–186.
\newblock \url{https://doi.org/10.1038/nature14279}

\bibitem[Li et~al., 2024]{li2024reconstructing}
Li, Z., Yang, W., Wu, P., Shan, Y., Zhang, X., Chen, F., Yang, J., \& Yang,
  J.-R. (2024).
\newblock Reconstructing cell lineage trees with genomic barcoding: approaches
  and applications.
\newblock {\em Journal of Genetics and Genomics}, 51(1), 35–47.
\newblock \url{https://doi.org/10.1016/j.jgg.2023.05.011}

\bibitem[Lie, 2023]{lie2023cerebras}
Lie, S. (2023).
\newblock Cerebras architecture deep dive: First look inside the
  hardware/software co-design for deep learning.
\newblock {\em IEEE Micro}, 43(3), 18–30.
\newblock \url{https://doi.org/10.1109/mm.2023.3256384}

\bibitem[McKinney, 2010]{mckinney2010data}
McKinney, W. (2010).
\newblock Data structures for statistical computing in python.
\newblock {\em Proceedings of the 9th Python in Science Conference}, SciPy,
  56–61.
\newblock \url{https://doi.org/10.25080/majora-92bf1922-00a}

\bibitem[Miller et~al., 2010]{miller2010creating}
Miller, M.~A., Pfeiffer, W., \& Schwartz, T. (2010).
\newblock Creating the cipres science gateway for inference of large
  phylogenetic trees.
\newblock {\em 2010 Gateway Computing Environments Workshop (GCE)}.
\newblock \url{https://doi.org/10.1109/gce.2010.5676129}

\bibitem[Moreno \& Dolson, 2024]{moreno2024hstrat}
Moreno, M.~A. \& Dolson, E. (2024).
\newblock {\em mmore500/hstrat}.
\newblock \url{https://doi.org/10.5281/zenodo.10779245}

\bibitem[Moreno et~al., 2022a]{moreno2022hereditary}
Moreno, M.~A., Dolson, E., \& Ofria, C. (2022a).
\newblock Hereditary stratigraphy: Genome annotations to enable phylogenetic
  inference over distributed populations.
\newblock {\em The 2022 Conference on Artificial Life}, Alife 2022, 64.
\newblock \url{https://doi.org/10.1162/isal_a_00550}

\bibitem[Moreno et~al., 2022b]{moreno2022hstrat}
Moreno, M.~A., Dolson, E., \& Ofria, C. (2022b).
\newblock hstrat: a python package for phylogenetic inference on distributed
  digital evolution populations.
\newblock {\em Journal of Open Source Software}, 7(80), 4866.
\newblock \url{https://doi.org/10.21105/joss.04866}

\bibitem[Moreno et~al., 2023]{moreno2023toward}
Moreno, M.~A., Dolson, E., \& Rodriguez-Papa, S. (2023).
\newblock Toward phylogenetic inference of evolutionary dynamics at scale.
\newblock {\em The 2023 Conference on Artificial Life}, Alife 2023, 568--668.
\newblock \url{https://doi.org/10.1162/isal_a_00694}

\bibitem[Moreno et~al., 2025a]{moreno_2025_16898849}
Moreno, M.~A., Dolson, E., Zaman, L., Singhvi, V., Yang, C., \& Rodriguez~Papa,
  S. (2025a).
\newblock {\em hstrat}.
\newblock \url{https://doi.org/10.5281/zenodo.16898849}

\bibitem[Moreno et~al., 2025b]{moreno2025testing}
Moreno, M.~A., Ranjan, A., Dolson, E., \& Zaman, L. (2025b).
\newblock Testing the inference accuracy of accelerator-friendly approximate
  phylogeny tracking.
\newblock {\em 2025 IEEE Symposium on Computational Intelligence in Artificial
  Life and Cooperative Intelligent Systems (ALIFE-CIS)}, 1–9.
\newblock \url{https://doi.org/10.1109/alife-cis64968.2025.10979833}

\bibitem[Moreno et~al., 2024a]{moreno2024analysis}
Moreno, M.~A., {Rodriguez Papa}, S., \& Dolson, E. (2024a).
\newblock {\em Analysis of phylogeny tracking algorithms for serial and
  multiprocess applications}.
\newblock \url{https://doi.org/10.48550/arXiv.2403.00246}

\bibitem[Moreno et~al., 2025c]{matthew_andres_moreno_2025_16898918}
Moreno, M.~A., Singhvi, V., Wagner, J., Dolson, E., \& Zaman, L. (2025c).
\newblock {\em mmore500/hstrat-reconstruction-algo: v1.0.1}.
\newblock \url{https://doi.org/10.5281/zenodo.16898918}

\bibitem[Moreno et~al., 2024b]{moreno2024trackable}
Moreno, M.~A., Yang, C., Dolson, E., \& Zaman, L. (2024b).
\newblock Trackable agent-based evolution models at wafer scale.
\newblock {\em The 2024 Conference on Artificial Life}, 87--98.
\newblock \url{https://doi.org/10.1162/isal_a_00830}

\bibitem[Moreno et~al., 2025d]{moreno_2025_16898904}
Moreno, M.~A., Yang, C., Dolson, E., \& Zaman, L. (2025d).
\newblock {\em async-ga: agent-based evolution on the cerebras wafer-scale
  engine (wse)}.
\newblock \url{https://doi.org/10.5281/zenodo.16898904}

\bibitem[Moreno et~al., 2024c]{moreno2024structured}
Moreno, M.~A., Zaman, L., \& Dolson, E. (2024c).
\newblock {\em Structured downsampling for fast, memory-efficient curation of
  online data streams}.
\newblock \url{https://doi.org/10.48550/arXiv.2409.06199}

\bibitem[Moshiri, 2020]{moshiri2020treeswift}
Moshiri, N. (2020).
\newblock Treeswift: A massively scalable python tree package.
\newblock {\em SoftwareX}, 11, 100436.
\newblock \url{https://doi.org/10.1016/j.softx.2020.100436}

\bibitem[Moshiri, 2025]{moshiri2025compacttree}
Moshiri, N. (2025).
\newblock Compacttree: a lightweight header-only c++ library and python wrapper
  for ultra-large phylogenetics.
\newblock {\em Gigabyte}, 2025.
\newblock \url{https://doi.org/10.46471/gigabyte.152}

\bibitem[Ofria \& Wilke, 2004]{ofria2004avida}
Ofria, C. \& Wilke, C.~O. (2004).
\newblock Avida: A software platform for research in computational evolutionary
  biology.
\newblock {\em Artificial Life}, 10(2), 191–229.
\newblock \url{https://doi.org/10.1162/106454604773563612}

\bibitem[pandas developers, 2020]{reback2020pandas}
pandas developers (2020).
\newblock pandas-dev/pandas: Pandas.
\newblock {\em Zenodo}.
\newblock \url{https://doi.org/10.5281/zenodo.3509134}

\bibitem[Pennock, 2007]{pennock2007models}
Pennock, R.~T. (2007).
\newblock Models, simulations, instantiations, and evidence: the case of
  digital evolution.
\newblock {\em Journal of Experimental \& Theoretical Artificial Intelligence},
  19(1), 29--42.

\bibitem[Rocklin, 2015]{rocklin2015dask}
Rocklin, M. (2015).
\newblock Dask: Parallel computation with blocked algorithms and task
  scheduling.
\newblock {\em Proceedings of the 14th Python in Science Conference}, SciPy,
  126–132.
\newblock \url{https://doi.org/10.25080/majora-7b98e3ed-013}

\bibitem[Sanderson, 2022]{sanderson2022taxonium}
Sanderson, T. (2022).
\newblock Taxonium, a web-based tool for exploring large phylogenetic trees.
\newblock {\em eLife}, 11.
\newblock \url{https://doi.org/10.7554/elife.82392}

\bibitem[Singhvi et~al., 2025]{supplemental}
Singhvi, V., Wagner, J., Dolson, E., Zaman, L., \& Moreno, M.~A. (2025).
\newblock {\em A scalable trie building algorithm for high-throughput
  phyloanalysis of wafer-scale digital evolution experiments}.
\newblock \url{https://doi.org/10.17605/osf.io/63ucz}

\bibitem[Stadler, 2013]{stadler2013recovering}
Stadler, T. (2013).
\newblock Recovering speciation and extinction dynamics based on phylogenies.
\newblock {\em Journal of Evolutionary Biology}, 26(6), 1203–1219.
\newblock \url{https://doi.org/10.1111/jeb.12139}

\bibitem[Stroud \& Ratcliff, 2025]{Stroud2025}
Stroud, J.~T. \& Ratcliff, W.~C. (2025).
\newblock Long-term studies provide unique insights into evolution.
\newblock {\em Nature}, 639(8055), 589–601.
\newblock \url{https://doi.org/10.1038/s41586-025-08597-9}

\bibitem[Vink et~al., 2024]{vink2024polars}
Vink, R., de~Gooijer, S., Beedie, A., Gorelli, M.~E., nameexhaustion, Peters,
  O., Guo, W., Burghoorn, G., van Zundert, J., Hulselmans, G., Grinstead, C.,
  Marshall, chielP, Turner-Trauring, I., Mitchell, L., Santamaria, M., Heres,
  D., Magarick, J., Genockey, K., ibENPC, deanm0000, Harbeck, H., Wilksch, M.,
  eitsupi, Koutsouris, I., Leitao, J., van Gelderen, M., Brannigan, L., \&
  Barbagiannis, P. (2024).
\newblock {\em pola-rs/polars: Python polars 1.16.0}.
\newblock \url{https://doi.org/10.5281/zenodo.14244124}

\bibitem[Virtanen et~al., 2020]{2020SciPy-NMeth}
Virtanen, P., Gommers, R., Oliphant, T.~E., Haberland, M., Reddy, T.,
  Cournapeau, D., Burovski, E., Peterson, P., Weckesser, W., Bright, J., {van
  der Walt}, S.~J., Brett, M., Wilson, J., Millman, K.~J., Mayorov, N., Nelson,
  A. R.~J., Jones, E., Kern, R., Larson, E., Carey, C.~J., Polat, {\.I}., Feng,
  Y., Moore, E.~W., {VanderPlas}, J., Laxalde, D., Perktold, J., Cimrman, R.,
  Henriksen, I., Quintero, E.~A., Harris, C.~R., Archibald, A.~M., Ribeiro,
  A.~H., Pedregosa, F., {van Mulbregt}, P., \& {SciPy 1.0 Contributors} (2020).
\newblock {{SciPy} 1.0: Fundamental Algorithms for Scientific Computing in
  Python}.
\newblock {\em Nature Methods}, 17, 261--272.
\newblock \url{https://doi.org/10.1038/s41592-019-0686-2}

\bibitem[Wang et~al., 2020]{wang2020role}
Wang, J.-T., Lin, Y.-Y., Chang, S.-Y., Yeh, S.-H., Hu, B.-H., Chen, P.-J., \&
  Chang, S.-C. (2020).
\newblock The role of phylogenetic analysis in clarifying the infection source
  of a covid-19 patient.
\newblock {\em Journal of Infection}, 81(1), 147–178.
\newblock \url{https://doi.org/10.1016/j.jinf.2020.03.031}

\bibitem[Waskom, 2021]{waskom2021seaborn}
Waskom, M.~L. (2021).
\newblock seaborn: statistical data visualization.
\newblock {\em Journal of Open Source Software}, 6(60), 3021.
\newblock \url{https://doi.org/10.21105/joss.03021}

\bibitem[Wiser et~al., 2013]{wiser2013long}
Wiser, M.~J., Ribeck, N., \& Lenski, R.~E. (2013).
\newblock Long-term dynamics of adaptation in asexual populations.
\newblock {\em Science}, 342(6164), 1364--1367.

\bibitem[Yang et~al., 2025]{yang2025downstream}
Yang, C., Wagner, J., Dolson, E., Zaman, L., \& Moreno, M.~A. (2025).
\newblock {\em Downstream: efficient cross-platform algorithms for
  fixed-capacity stream downsampling}.
\newblock \url{https://doi.org/10.48550/arXiv.2506.12975}

\end{thebibliography}


\begin{thebibliography}{}

\bibitem[Davies, 1987]{davies1987hypothesis}
Davies, R.~B. (1987).
\newblock Hypothesis testing when a nuisance parameter is present only under
  the alternative.
\newblock {\em Biometrika}, 74(1), 33--43.

\bibitem[Jakob et~al., 2017]{wenzel2017pybind11}
Jakob, W., Rhinelander, J., \& Moldovan, D. (2017).
\newblock {\em pybind11 — seamless operability between c++11 and python}.
\newblock https://github.com/pybind/pybind11.

\bibitem[Moreno, 2024]{moreno2024colorclade}
Moreno, M.~A. (2024).
\newblock {\em mmore500/colorclade}.
\newblock \url{https://doi.org/10.5281/zenodo.10802404}

\bibitem[Moreno et~al., 2022]{moreno2022hstrat}
Moreno, M.~A., Dolson, E., \& Ofria, C. (2022).
\newblock hstrat: a python package for phylogenetic inference on distributed
  digital evolution populations.
\newblock {\em Journal of Open Source Software}, 7(80), 4866.
\newblock \url{https://doi.org/10.21105/joss.04866}

\bibitem[Moreno et~al., 2024]{moreno2024analysis}
Moreno, M.~A., {Rodriguez Papa}, S., \& Dolson, E. (2024).
\newblock {\em Analysis of phylogeny tracking algorithms for serial and
  multiprocess applications}.
\newblock \url{https://doi.org/10.48550/arXiv.2403.00246}

\bibitem[Pilgrim, 2021]{pilgrim2021piecewise}
Pilgrim, C. (2021).
\newblock piecewise-regression (aka segmented regression) in python.
\newblock {\em Journal of Open Source Software}, 6(68), 3859.
\newblock \url{https://doi.org/10.21105/joss.03859}

\bibitem[Singhvi et~al., 2025]{supplemental}
Singhvi, V., Wagner, J., Dolson, E., Zaman, L., \& Moreno, M.~A. (2025).
\newblock {\em A scalable trie building algorithm for high-throughput
  phyloanalysis of wafer-scale digital evolution experiments}.
\newblock \url{https://doi.org/10.17605/osf.io/63ucz}

\end{thebibliography}
\end{document}